\documentclass[11pt,a4paper]{article}
\usepackage{times,latexsym}
\usepackage{url}
\usepackage[T1]{fontenc}
\usepackage{amsmath}
\usepackage{amsfonts}
\usepackage{bm}
\usepackage{graphicx}
\usepackage{dsfont}
\usepackage{booktabs} 
\usepackage{multirow}
\usepackage{pbox}
\usepackage{array}
\usepackage{adjustbox}
\usepackage{hyperref}
\usepackage{multicol}
\usepackage{algorithm}
\usepackage{algpseudocode}
\usepackage{algorithmicx}

\usepackage[acceptedWithA]{tacl2021v1}
\usepackage{xspace,mfirstuc,tabulary}

\newif\iftaclinstructions
\taclinstructionsfalse 
\iftaclinstructions
\renewcommand{\confidential}{}
\renewcommand{\anonsubtext}{(No author info supplied here, for consistency with
TACL-submission anonymization requirements)}
\newcommand{\instr}
\fi

\iftaclpubformat 

\else

\fi

\title{LLM Reading Tea Leaves: Automatically Evaluating Topic Models with Large Language Models}

\author{
  Xiaohao Yang$^\diamond$ 
  \
  He Zhao$^\dagger$\Thanks{Corresponding authors: Lan Du, He Zhao}
  \
    Dinh Phung$^\diamond$ 
\ Wray Buntine$^\ddag$
  \
  Lan Du$^{\diamond\ast}$
  \\
  \ \\
  $^\diamond$Monash University \\Melbourne, Australia\\
  \texttt{\{xiaohao.yang,dinh.phung,lan.du\}@monash.edu}
  \\
  \\
  $^\dagger$CSIRO’s Data61\\
  Sydney, Australia\\
  \texttt{he.zhao@data61.csiro.au,}
  \\
  \\
$^\ddag$VinUniversity\\
Hanoi, Vietnam\\
\texttt{wray.b@vinuni.edu.vn,}  
}

\date{}

\begin{document}
\maketitle
\begin{abstract}
Topic modeling has been a widely used tool for unsupervised text analysis. However, comprehensive evaluations of a topic model remain challenging. Existing evaluation methods are either less comparable across different models (e.g., perplexity) or focus on only one specific aspect of a model (e.g., topic quality or document representation quality) at a time, which is insufficient to reflect the overall model performance.
In this paper, we propose WALM (\textbf{W}ord \textbf{A}greement with \textbf{L}anguage \textbf{M}odel), a new evaluation method for topic modeling that considers the semantic quality of document representations and topics in a joint manner, leveraging the power of Large Language Models (LLMs). With extensive experiments involving different types of topic models, WALM is shown to align with human judgment and can serve as a complementary evaluation method to the existing ones, bringing a new perspective to topic modeling. Our software package is available at \href{https://github.com/Xiaohao-Yang/Topic\_Model\_Evaluation}{https://github.com/Xiaohao-Yang/Topic\_Model\_Evaluation}.

  
\end{abstract}

\section{Introduction}
Topic modeling~\cite{blei2003latent}, a popular unsupervised text analysis technique, has been applied to various domains, including information retrieval \citep{yi2009comparative}, marketing analysis \citep{reisenbichler2019topic}, social media analysis \citep{laureate2023systematic}, bioinformatics \citep{liu2016overview}, and more. A topic model typically learns a set of global topics to interpret a text corpus and the topic proportion of a document as its semantic representation. 

Although topic models have been time-tested for two decades, as an unsupervised technique, comprehensive evaluations of a topic model remain challenging~\citep{ijcai2021p638}. Originally, topic models are implemented as probabilistic graphical models such as Latent Dirichlet Allocation (LDA) \citep{blei2003latent} and many of its Bayesian extensions, e.g., in~\citet{10.1145/1667053.1667056,paisley2015nested,gan2015learning,zhou2016augmentable,zhao2018inter,zhao2018dirichlet}. For these models, it has been common practice to measure the log-likelihood or perplexity of a model on held-out test documents. While log-likelihood or perplexity provides a straightforward quantitative comparison between models, several issues still persist. Since topic models are not primarily designed to predict words in documents but rather to learn semantically meaningful topics and interpretable document representations, these metrics fail to capture these aspects. Furthermore, estimating the predictive probability is often intractable for Bayesian models, and different papers may employ different sampling or approximation techniques~\citep{wallach2009evaluation,buntine2009estimating}. For recently proposed Neural Topic Models (NTMs)~\citep{ijcai2021p638}, the computation of log-likelihood is even more inconsistent.

In addition to log-likelihood or perplexity, document representation quality and topic quality are evaluated separately. For document representation quality, downstream task performance is typically used as a metric, such as document classification~\citep{yang2023towards}, clustering~\citep{ijcai2021p638}, and retrieval~\citep{larochelle2012neural}. For topic quality, the ultimate evaluation method is human evaluation, which is time-consuming and expensive. Thus, various automatic metrics have been proposed, such as topic coherence~\citep{lau2014machine}, which measures how semantically coherent the representative words in a topic are, and topic diversity~\citep{dieng-etal-2020-topic}, which measures how diverse discovered topics are. To comprehensively evaluate the performance of a topic model, one needs to report multiple metrics on both document representation and topic qualities. However, these metrics can be contradictory, e.g., a topic model with good topic quality may not preserve good quality on document representation, and vice versa. This discrepancy complicates the model selection process for topic models in practice.

In this paper, we aim to develop a new evaluation approach for topic modeling that considers both the semantic quality of document representations and topics in a joint manner, leveraging the power of Large Language Models (LLMs). Our key idea is as follows: After being trained, a topic model can infer a document's distribution over topics and each topic is a distribution over vocabulary words. With these two distributions, a model can generate a set of ``topical'' words given a document, such as by looking at its representative topics and the representative words of each topic. The generation of the topical words takes both the topic distribution of a document and the word distributions of the topics into account, which captures the semantic summary of the document and is expected to align with the keywords identified by humans.
Given the high cost of human evaluation, we propose using LLMs as a proxy by employing appropriate prompts to generate keywords for the document, which are then compared with the topical words produced by a topic model. Finally, to quantify the agreement between the words from the topic model and the LLM, a series of WALM (\textbf{W}ord \textbf{A}greement with \textbf{L}anguage \textbf{M}odel) metrics are proposed. WALM has the following appealing properties:
\begin{itemize}
    \item It is a joint metric that evaluates the quality of both document representations and topics. 
    \item It assesses how effectively a topic model captures the semantics of a document, which is a core objective of topic modeling.
    \item It allows for comparisons across various types of topic models.
\end{itemize}
To examine WALM series metrics, we conduct extensive experiments using various popular topic models on different datasets, comparing them with other widely used topic model evaluation metrics. Moreover, human evaluation is also conducted to demonstrate the alignment of WALM with human judgment.

\section{Related Work}
As an unsupervised technique for uncovering hidden themes in text, evaluating topic models remains challenging. Early evaluations of a topic model rely on the log-likelihood or perplexity of held-out documents \citep{blei2003latent}, which measures how well the model predicts the words of documents. As the computation of predictive probability is often intractable for conventional Bayesian topic models, various sampling or approximation techniques have been proposed \cite{wallach2009evaluation,buntine2009estimating}. Apart from the inconsistent estimation, held-out likelihood is regarded as not correlated with the interpretability of topics from a human perspective \citep{chang2009reading}, prompting the direct evaluation of topics and document representation quality.

As for the evaluation of topics, \citet{chang2009reading} design the word and topic intrusion tasks for human annotators, where high-quality topics or document representations are those where annotators can easily identify the intruders. \citet{newman2010automatic} and \citet{mimno2011optimizing} evaluate topic coherence by direct ratings from human experts. Although human judgment is commonly regarded as the gold standard, it is expensive and impractical for large-scale evaluation. Automated evaluation of topic coherence is more practical, such as Normalized Pointwise Mutual Information (NPMI) \citep{lau2014machine}, which relies on the co-occurrence of the topic's top words in the reference corpus to measure topic coherence, with the underlying assumption that a large reference corpus such as Wikipedia can capture prevalent language patterns. Although they automate the evaluation of topics and strongly correlate with human judgment \citep{newman2010automatic}, counting word co-occurrence in a large reference corpus is still relatively expensive. Moreover, coherence metrics can vary depending on the reference corpus, and there is no single ``right'' reference corpus that is suitable for all datasets \cite{doogan2021topic}. Recent works propose leveraging word embeddings \citep{nikolenko2016topic} or contextualized embeddings \citep{hoover-etal-2021-linguistic} for efficiently evaluating topic coherence, incorporating semantics from pre-trained embeddings. Due to common posterior collapse issues \citep{lucas2019don} in the growing field of neural topic models \citep{ijcai2021p638}, recent works also consider topic diversity \citep{dieng-etal-2020-topic} during evaluation, which measures how distinct the top words of each topic are.

As for the evaluation of document representation, early works focus on how well the topic proportion of a document represents the document content, assessed through a topic intrusion task by human annotators \citep{chang2009reading}, which is further extended as automated metrics \citep{bhatia-etal-2017-automatic,bhatia-etal-2018-topic}. Recent topic models often use the topic proportions as document representations, the quality of which is commonly investigated through downstream tasks, including their use as features for document classification \citep{nguyen2021contrastive}, clustering \citep{zhao2020neural}, and retrieval \citep{larochelle2012neural}. Recently, the generalization ability of topic models is investigated by evaluating their quality of document representations across different unseen corpora \citep{yang2023towards}.

In the era of Large Language Models (LLMs) \citep{brown2020language,thoppilan2022lamda,touvron2023llama,touvron2023llama2,10.5555/3648699.3648939}, recent research has begun leveraging LLMs to evaluate topic models, such as using ChatGPT \footnote{https://openai.com/index/chatgpt/} as a proxy for human annotators for word intrusion and topic rating tasks for evaluating topic coherence \citep{stammbach-etal-2023-revisiting,rahimi-etal-2024-contextualized}. The focus of these works is still on topic quality only.

In this work, we propose new evaluation metrics for topic models, differing from previous works in the following ways: (1) Unlike evaluations that focus on only sub-components of a topic model (i.e., topics or document representations), our evaluation metrics offer a joint approach to topic model evaluation, considering both topics and document representations together. (2) Compared with log-likelihood or perplexity, which also evaluate based on documents, our evaluation metrics consider semantics from documents and align with the focus of topic modeling. (3) Different from recent LLM-based evaluations that use LLMs for topic quality evaluation, ours considers both topic quality and document representation quality and our use of LLMs is quite different from previous works.

\begin{figure*}[!ht]
    \centering
\includegraphics[width=0.85\textwidth,height=\textheight,keepaspectratio]{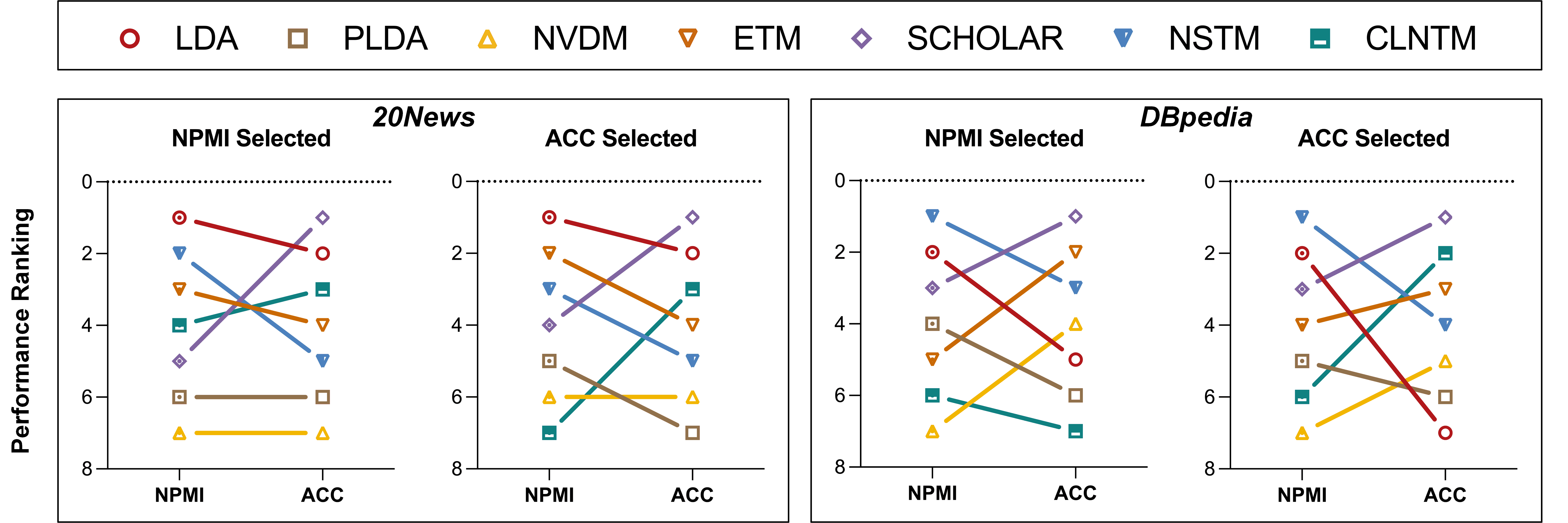}
    \caption{Performance rankings of topic quality (NPMI) and document representation quality (ACC) during model selection. The best model state/checkpoint can be determined using either NPMI or ACC as the selection criterion. However, it can be observed that the rankings for topic quality and document representation quality are inconsistent under the same selection criteria. Experiments are conducted five times, with the number of topics set to 50.}
    \label{model_ranking}
\end{figure*}

\section{Background}
Given a document collection $\mathcal{D}:=\{\bm{d}_1,...,\bm{d}_M\}$ with $V$ vocabulary words, a topic model is typically trained on their Bag-of-Words (BOWs), e.g., $\bm{x}\in\mathbb{N}^{V}$. The topic model can infer a distribution over $K$ topics for each document by running its inference process: 
\begin{equation}\label{eq:inf}
    \bm{z}:=f_{\bm{\theta}}(\bm{x}),
\end{equation}
where $\bm{\theta}$ denotes the model parameters of the inference process; $\bm{z}\in\Delta^{K}$ ($\Delta$ denotes the probability simplex) indicates the proportion of each topic present in the document and is commonly used as its semantic representation. Besides, the topic model also discovers $K$ global topics for the corpus (i.e., $\mathcal{T}:=\{\bm{t}_1,...,\bm{t}_K\}$), where each topic $\bm{t}\in\Delta^{V}$ is a distribution over $V$ vocabularies. Ideally, each topic captures a semantic concept that can be interpreted by its top-weighted words. To train a topic model, one often needs to generate or reconstruct the word distribution of the document from $\bm{z}$ by running its generative process:
\begin{equation}\label{eq:gen}
    \bm{w}:=f_{\bm{\phi}}(\bm{z}, \mathcal{T}),
\end{equation}
where $\bm{\phi}$ are the model parameters of the generative process; $\bm{w}\in\Delta^{V}$ is the per-document word distribution from which $\bm{x}$ is sampled. Let $\mathcal{Z}:=\{\bm{z}_1,...,\bm{z}_N\}$ be the semantic representations of $N$ test documents and $\mathcal{T}$ be the $K$ learned topics, current evaluation of a topic model is commonly conducted based on either $\mathcal{Z}$ or $\mathcal{T}$ separately.

\section{Method}
\subsection{Motivation}
Both topics and document representations are important components of a topic model. To comprehensively evaluate a topic model, it is common practice to report the performance of both parts. This can be done by measuring topic quality using metrics such as NPMI and assessing document representation quality through downstream classification accuracy (ACC) (see section \ref{existing_setting} for details of metrics calculation). However, a model that prioritizes topic quality (e.g., NPMI) may not perform well in terms of document representations (e.g., ACC), and vice versa, which creates difficulty during model selection, as illustrated in Figure \ref{model_ranking}. This inconsistency in the performance of the two components is also indicated by \citet{bhatia-etal-2017-automatic}. Therefore, evaluating a topic model based on  sub-components only is insufficient to reveal the entire model's performance. Recent topic models often focus on improving topic quality, such as clustering-based models \citep{sia-etal-2020-tired,grootendorst2022bertopic}, but they do not evaluate their effectiveness in representing documents.
In this work, we aim to introduce a novel evaluation method for topic modeling that jointly assesses the semantic quality of both topics and document representations, with the help of large language models.

\subsection{Key Idea}
We propose to conduct the evaluation in a joint manner that considers both document representations and topics, rather than evaluating them separately as in previous works. To do so, we obtain the document-word distribution $\bm{w}$ for a given document from the topic model by running both its inference and generative process: 
\begin{equation}\label{word_dist}
\bm{w} := f_{\bm{\phi}}(f_{\bm{\theta}}(\bm{x})).    
\end{equation}
The inference process infers the document representation 
$\bm{z}$ for a given document $\bm{x}$, as in Eq. \ref{eq:inf}; the generative process\footnote{We omit topics $\mathcal{T}$ in Eq. \ref{word_dist} as they are considered part of the parameters of the generative process $\bm{\phi}$.} generates or reconstructs the word distribution $\bm{w}$ based on the document representation 
$\bm{z}$ and topics $\mathcal{T}$, as in Eq. \ref{eq:gen}. Therefore, the evaluation based on $\bm{w}$ involves both document representations and topics.

Next, we take the top-weighted words 
$\mathbf{w}$ from the word distribution 
$\bm{w}$ generated by the topic model as the ``topical'' words of the document. Those topical words can be regarded as a semantic summary of the document from the target topic model's perspective. To generate high-quality topical words for a document, a topic model should learn good global topics as well as good document representations.
Now, the evaluation of a topic model can be reframed as assessing the quality of its topical words. Suppose the true representative words $\mathbf{k}$ of document $\bm{x}$ are given, then we can formulate our evaluation task as:
\begin{equation} \label{eq1}
S(\mathbf{w}, \mathbf{k}),
\end{equation}
where $S(\cdot,\cdot)$ is a score function (Sec. \ref{sfunction}) to quantify the agreement between $\mathbf{w}$ and $\mathbf{k}$ (Sec. \ref{truth}).

\begin{figure}[!t]
    \centering
\includegraphics[width=0.48\textwidth,height=\textheight,keepaspectratio]{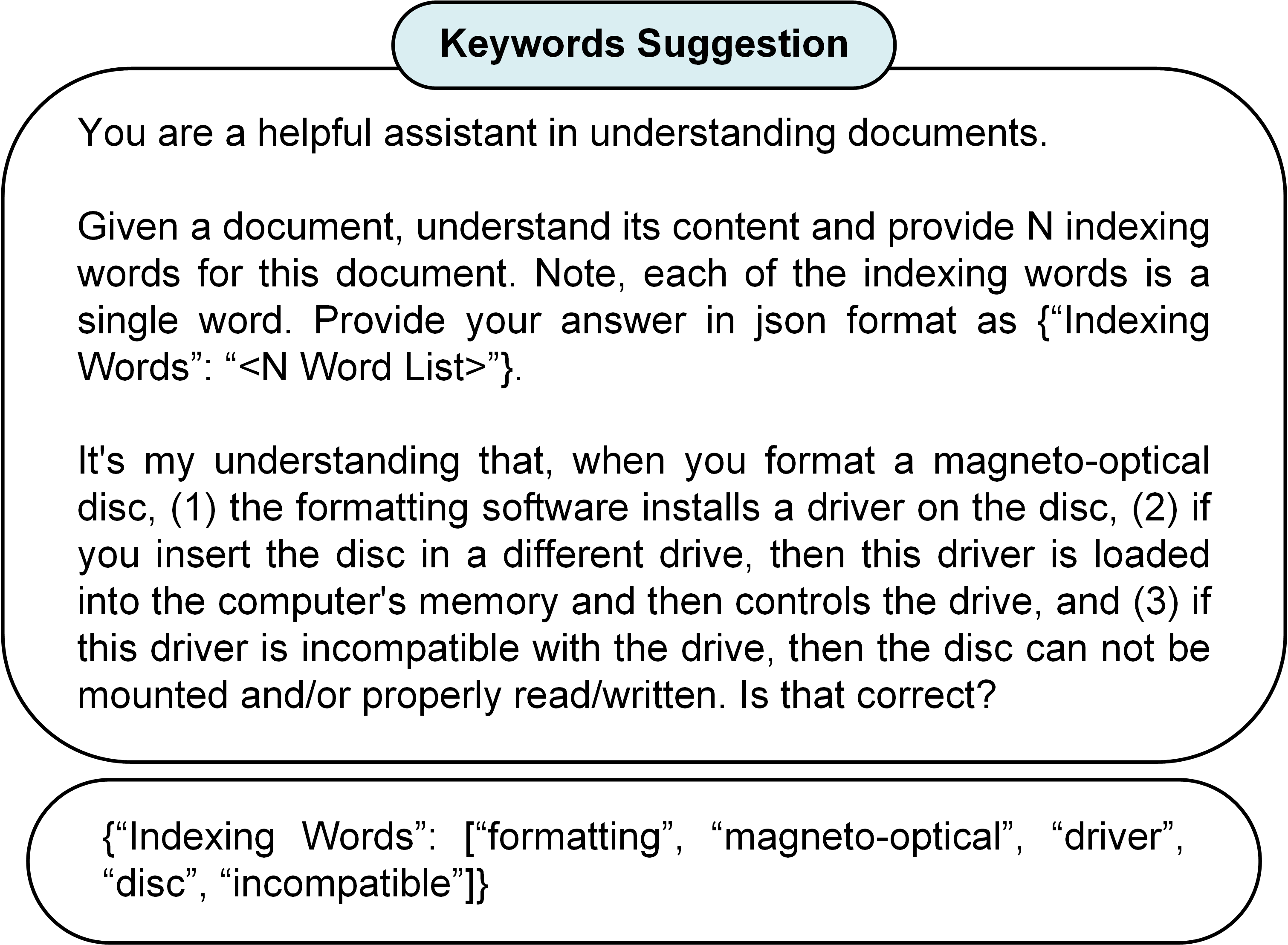}
\vspace{-5mm}
    \caption{An example prompt and output of keywords suggestion by the LLM. In this example, the number of keywords (i.e., N) is set to 5.}
    \label{fig:prompt_kw}
\end{figure}

\subsection{Word Suggestion by LLM}\label{truth}
\paragraph{Keywords Suggestion}
Following our evaluation task in Eq. \ref{eq1}, the ideal representative words $\mathbf{k}$ are from human
summary of the document. However, this is expensive and impractical for large-scale evaluation. With the recent advancements in LLMs, which have demonstrated performance akin to human capabilities in various natural language processing tasks, including text summarization \cite{wang-etal-2023-zero,tang2023evaluating,zhang2024benchmarking} and keyphrase extraction \cite{song2023large,maragheh2023llm,10.1145/3627673.3680093}, we propose leveraging LLMs through prompting to generate keyword suggestions for a given document:
\begin{equation}\label{eq:prompt}
\mathbf{k}:=\text{LLM}(\text{Prompt($\bm{d}$)}).
\end{equation}
Specifically, we query keywords $\mathbf{k}$ for a given document $\bm{d}$ from an LLM by proposing the prompt shown in Figure \ref{fig:prompt_kw}. The prompt consists of a task instruction and the queried document. 

\paragraph{Topic-Aware Keywords Suggestion}
Analogous to the generation of topical words in topic modeling -- where global topics of the document collections are identified first, followed by associated keywords for each given document -- we propose prompting the LLM in a similar manner, ensuring it considers collection-level topics when providing keywords suggestion for each document, written as:
\begin{equation}\label{eq:prompt_new}
\mathbf{k}:=\text{LLM}(\text{Prompt($\bm{d},\mathcal{T}$)}),
\end{equation}
where $\mathcal{T}$ denotes the set of topics of the text corpus. To obtain the collection-level topics $\mathcal{T}$ for the corpus by the LLM, we follow the topic generation approach by \citet{pham-etal-2024-topicgpt}. Briefly, it leverages an LLM to iteratively identify new topics from each document. A subsequent refinement process then merges similar topics and removes those with low frequency. For further details on the topic generation process, we refer readers to \citet{pham-etal-2024-topicgpt} (section 3.1). 

\begin{figure}[!t]
    \centering
\includegraphics[width=0.48\textwidth,height=\textheight,keepaspectratio]{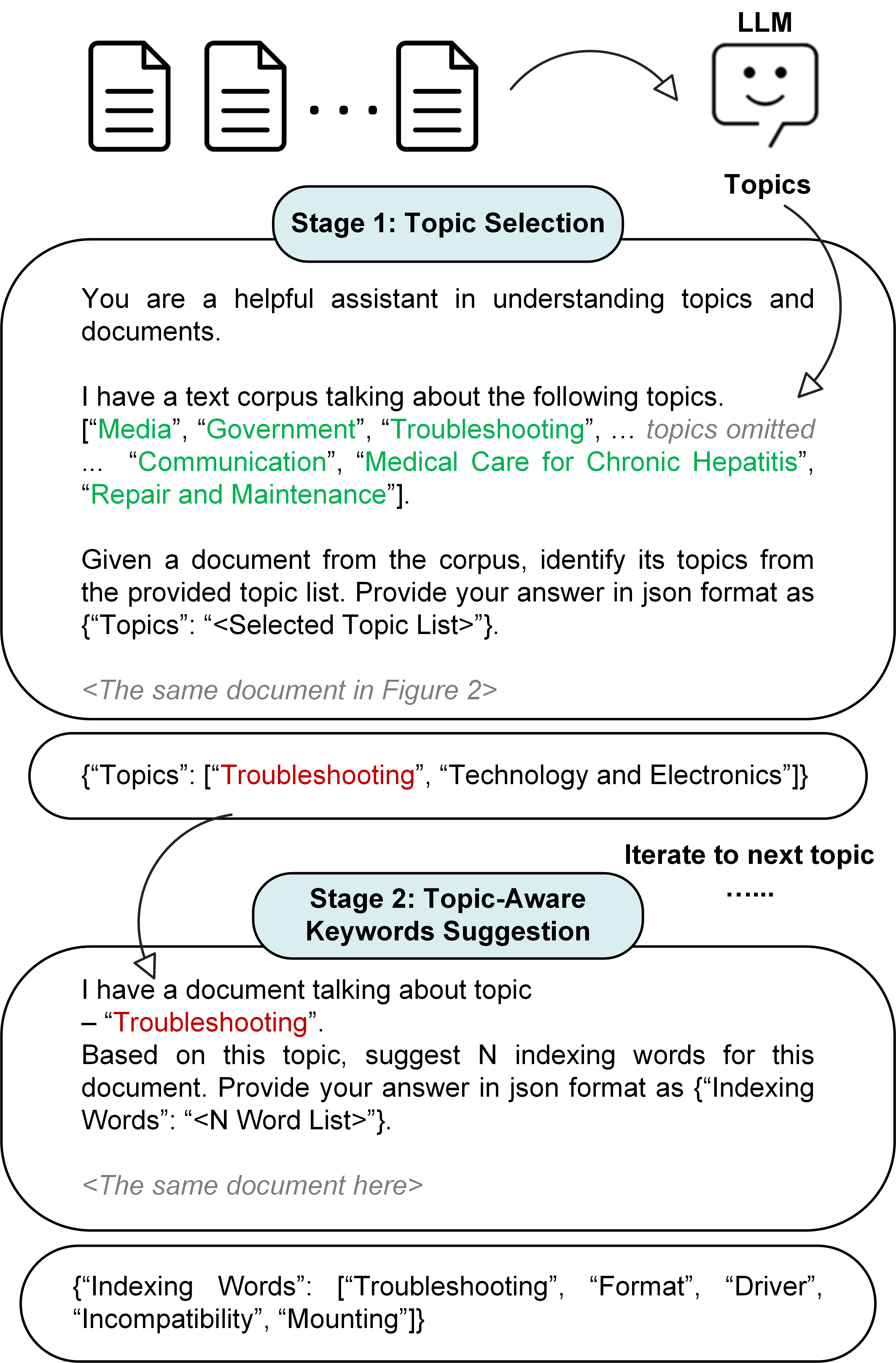}
\vspace{-5mm}
    \caption{An illustration of topic-aware keywords suggestion pipeline. The words highlighted in green represent collection-level topics generated by the LLM. Each topic selected in stage 1 is used in the stage 2 prompt to generate topic-aware keywords.}
    \label{fig:prompt_kw_topic-aware}
\end{figure}

Using these corpus-level topics, we prompt the LLM to generate keywords for each document in a two-stage process, considering the overarching themes of the collection. In the first stage, the LLM selects relevant topics for the target document from the corpus-level topics. In the second stage, we prompt the LLM to generate indexing words for the document based on each selected topic. The final set of keywords is obtained by merging the words generated for each selected topic. An example prompt and output for topic-aware keywords suggestion is shown in Figure \ref{fig:prompt_kw_topic-aware}.

\subsection{Choices of the Score Function}\label{sfunction}
For the score function $S(\cdot,\cdot)$ in Eq. \ref{eq1}, we propose different ways to calculate it: (1) Overlap-based, which computes the number of overlapping words between $\mathbf{w}$ and $\mathbf{k}$, and (2) Embedding-based, which calculates the overall semantic similarity between the two word sets using pre-trained word embeddings.

\paragraph{Word Overlap} A straightforward choice of the score function is directly counting the overlaps between $\mathbf{w}$ and $\mathbf{k}$. Considering the potential variant in forms of the same word, we convert each word to its root form before counting, formulated as:
\begin{equation} \label{metric_overlap}
S_{\text{overlap}}:=C(f_{\text{root}}(\mathbf{w}) \cap f_{\text{root}}(\mathbf{k}))\times f_{\text{n}}(\mathbf{w},\mathbf{k}),
\end{equation}
where $C(\cdot)$ and $f_{\text{root}}(\cdot)$ are the counting and rooting (e.g., stemming or lemmatization) operation, respectively; $f_{\text{n}}(\mathbf{w},\mathbf{k}):=1/(N+M)$ returns the normalising factor based on two input word sets, where $N$ and $M$ are the number of words in $\mathbf{w}$ and $\mathbf{k}$, respectively.

\paragraph{Synset Overlap}
Considering the case that different words may describe the same or similar concept (e.g., ``puppy'' and ``dog''), we leverage WordNet \citep{miller1995wordnet} synsets to determine word overlaps: if the synsets of two words intersect, they are considered to overlap. Then, we define the synset overlap score as:
\begin{align*} 
S_{\text{synset}}  :=  (\sum_{i=1}^{N}\sum_{j=1}^{M} \mathds{1}(C(f_{\text{synset}}(\mathrm{w}_i) \cap \\ f_{\text{synset}}(\mathrm{k}_j))>0))  \nonumber \times f_{\text{n}}(\mathbf{w},\mathbf{k}), \tag{7} 
\label{metric_synset}
\end{align*}
where $\mathds{1}(\cdot)$ denotes the indicator function; $f_{\text{synset}}(\cdot)$ is a function that returns the synset for a given word. Intuitively, the synset-based score builds on the idea of word overlap, considering two words as overlapping if their synsets intersect, rather than requiring an exact match.

\paragraph{Word Optimal Assignment} 
We consider another choice of $S$, which measures the overall semantic similarity between two sets of words with pre-trained word embeddings \citep{mikolov2013efficient,pennington2014glove}. Since the alignments between words from $\mathbf{w}$ and $\mathbf{k}$ are unknown, directly measuring similarity between word embeddings is not feasible. To automatically find the alignment for each word of $\mathbf{w}$ to each word of $\mathbf{k}$, we formulate it as the following Optimal Assignment (OA) problem and solve it using the Hungarian algorithm \cite{kuhn1955hungarian}: Given a word set $\mathbf{w}$ that has $N$ words: $\mathbf{w}:=\{\mathrm{w}_1,\mathrm{w}_2,...,\mathrm{w}_{N}\}$ and their embedding vectors $\bm{E}^{\mathbf{w}}:=\{\bm{e}^{\mathrm{w}_1},\bm{e}^{\mathrm{w}_2},...,\bm{e}^{\mathrm{w}_{N}}\}$; and another word set $\mathbf{k}$ that has $M$ words: $\mathbf{k}:=\{\mathrm{k}_1,\mathrm{k}_2,...,\mathrm{k}_{M}\}$ with related embedding vectors $\bm{E}^{\mathbf{k}}:=\{\bm{e}^{\mathrm{k}_1},\bm{e}^{\mathrm{k}_2},...,\bm{e}^{\mathrm{k}_{M}}\}$. Define a cost matrix $\bm{C}\in\mathbb{R}_{\ge 0}^{N\times M}$ whose entry $C_{i,j}:=\text{CosD}(\bm{e}^{\mathrm{w}_i},\bm{e}^{\mathrm{k}_j})$, where $\text{CosD}(\cdot,\cdot)$ denotes the cosine distance function; and a binary matrix $\bm{A}\in \{0,1\}^{N\times M}$ whose entry $A_{i,j}=1$ if word $\mathrm{w}_i$ is assigned to word $\mathrm{k}_j$, and 0 otherwise. The goal is to solve the following optimal assignment problem:
\begin{align*}\label{oa}
\min_{\bm{A}}\sum_{i=1}^{N} & \sum_{j=1}^{M}C_{i,j}\times A_{i,j}, 
\tag{8}
\end{align*}
subject to $\sum_{j=1}^{M} A_{i,j} = 1$ and $\sum_{i=1}^{N} A_{i,j} = 1$.
By finding the optimal binary matrix $\bm{A}^{*}$, we obtain the distance between $\mathbf{w}$ and $\mathbf{k}$ by:
\begin{equation} \label{oa_score}
S_{\text{oa}}:=D_{\text{oa}}(\bm{\mathbf{w}},\bm{\mathbf{k}}):= \sum_{i=1}^{N}\sum_{j=1}^{M}C_{ij}\times A_{ij}^{*}. \tag{9}
\end{equation}

\paragraph{Word Optimal Transport}
Optimal Transport (OT) has recently been used as a powerful geometric tool to measure the distance between distributions, with rich applications in machine learning and related areas~\cite{ge2021ota,zhaoneural2021,nguyen2021most,wanrepresenting2022,danlearning2022,buiunified,vuong2023vector,zhao2023transformed,ye2024ptarl,vo2024optimal,gao2024distribution}.
Considering that both $\mathbf{w}$ and $\mathbf{k}$ are top words of probability distributions, where each word essentially retains a portion of probability mass. Our previous calculations ignore the probability mass of words and treat each word in the set as equal. Now we include the probability mass and formulate the similarity calculation between $\mathbf{w}$ and $\mathbf{k}$ as an Optimal Transport (OT) problem: Given two discrete distributions $\mu(\mathbf{w}, \bm{w})$ and $\mu(\mathbf{k}, \bm{k})$, where $\mathbf{w}:=\{\mathrm{w}_1,\mathrm{w}_2,...,\mathrm{w}_{N}\}$ and $\mathbf{k}:=\{\mathrm{k}_1,\mathrm{k}_2,...,\mathrm{k}_{M}\}$ are the supports of those two distributions; $\bm{w}\in\Delta^{N}$ and $\bm{k}\in\Delta^{M}$ are their related probability vectors\footnote{We assume $\mu(\mathbf{k}, \bm{k})$ is a uniform distribution over the keywords $\mathbf{k}$ from the LLM. Thus, $\bm{k}$ is a uniform probability vector.}; following the same construction of cost matrix $\bm{C}$ in the previous OA problem, the OT problem between $\mu(\mathbf{w}, \bm{w})$ and $\mu(\mathbf{k}, \bm{k})$ is defined as:
\begin{align*}\label{ot}
    \min_{\bm{P}} \sum_{i=1}^{N} & \sum_{j=1}^{M} C_{i,j}\times P_{i, j}, 
    \tag{10}
\end{align*}
subject to $\sum_{j=1}^{M} P_{i,j} = w_i$ and $\sum_{i=1}^{N} P_{i,j} = k_{j}$; $\bm{P}\in\mathbb{R}_{\ge 0}^{N\times M}$ is the transport plan, whose entry $P_{i,j}$ indicates the amount of probability mass moving from $w_i$ to $k_j$. Similarly, by finding the optimal transport plan $\bm{P}^{*}$ using solvers such as those in \citet{flamary2021pot}, the OT distance between $\mu(\mathbf{w}, \bm{w})$ and $\mu(\mathbf{k}, \bm{k})$ is obtained by:
\begin{align*}\label{ot_score}
S_{\text{ot}}& :=D_{\text{ot}}(\mu(\mathbf{w},\bm{w}),\mu(\mathbf{k}, \bm{k})) \\ & :=\sum_{i=1}^{N}\sum_{j=1}^{M}  C_{ij}\times P_{ij}^{*}. \tag{11}
\end{align*}

Compared with our OA and OT formulations for WALM, they are similar in that they both construct the cost matrix $\bm{C}$ using cosine distance between pre-trained word embeddings. However, they differ in the following ways: (1) OA treats words in the set as equal, while OT considers probability mass of each word. (2) OA can be viewed as a ``hard'' assignment problem between two word sets because the entries of 
$\bm{A}$ are binary. In contrast, OT can be regarded as a ``soft'' assignment because of the spread of probability mass in $\bm{P}$.

\section{Experiments}
\subsection{Experimental Setup}
\paragraph{Datasets} Two widely used datasets, 20Newsgroup \citep{lang1995newsweeder} (\textbf{20News}), which contains long documents, and \textbf{DBpedia} \citep{auer2007dbpedia}, which includes short documents, are used for our experiments. Our pre-processed datasets are available in the Github repository.


\paragraph{Evaluated Models} We conduct experiments on 7 popular topic models from traditional probabilistic to recent neural topic models. (1) Latent Dirichlet Allocation (\textbf{LDA}) \citep{blei2003latent}, the most popular probabilistic topic model that assumes a document is generated by a mixture of topics. (2) LDA with Products of Experts (\textbf{PLDA}) \citep{srivastava2017autoencoding}, an early NTM that applies the product of experts instead of the mixture of multinomials in LDA. (3) Neural Variational Document Model (\textbf{NVDM}) \citep{miao2017discovering}, a pioneer NTM that uses a Gaussian as the prior distribution of topic proportions of documents. (4) Embedded Topic Model (\textbf{ETM}) \citep{dieng-etal-2020-topic}, an NTM that involves word and topic embeddings in the generative process. (5) Neural Topic Model with Covariates, Supervision, and Sparsity (\textbf{SCHOLAR}) \citep{card-etal-2018-neural}, an NTM that applies a logistic normal prior on topic proportions and leverages extra information from metadata. (6) Neural Sinkhorn Topic Model (\textbf{NSTM}) \citep{zhao2020neural}, a recent NTM based on an optimal transport framework. (7) Contrastive Learning Neural Topic Model (\textbf{CLNTM}) \citep{nguyen2021contrastive}, a recent NTM that uses contrastive learning to regularize document representations. We keep all these models' default settings as suggested in their implementations. All experiments are conducted 5 times with different model random seeds; mean and standard deviation values are reported.

\paragraph{Settings of WALM} \label{our_setting} For the WALM settings, we use GloVe word embeddings pre-trained on Wikipedia \citep{pennington2014glove}\footnote{https://nlp.stanford.edu/projects/glove/} in our embedding-based metrics. For the LLM generation settings, we use LLAMA3-8B-Instruct\footnote{https://huggingface.co/meta-llama/Meta-Llama-3-8B-Instruct} for our main experiments. We employ greedy decoding during LLM generation to ensure deterministic outputs, setting the maximum number of generated tokens to 300. When prompting the LLM, we limit the number of generated keywords to 5. For topical words from the topic model, we select the top 10 weighted words from the document-word distribution for each given document.

\paragraph{Settings of Existing Metrics}\label{existing_setting}
We also evaluate topic models with existing commonly used metrics to compare with ours. (1) Topic Coherence and Diversity: we apply \textbf{NPMI} to evaluate topic coherence using Wikipedia as the reference corpus, done by the Palmetto package\footnote{https://github.com/dice-group/Palmetto} \citep{roder2015exploring}. Following standard protocol, we consider the top 10 words of each topic and obtain the average NPMI score of topics by selecting the top 50\% coherent topics. As for Topic Diversity (\textbf{TD}), we compute the percentage of unique words in the top 25 words of all topics, as defined in \citet{dieng-etal-2020-topic}. (2) Document Representation Quality: we conduct document classification and clustering to evaluate the representation capability of topic models. As for classification, we train a Random Forest classifier based on the training documents' representation and test the accuracy (\textbf{ACC}) in the testing documents, as in previous works such as \citet{nguyen2021contrastive}. As for clustering, we conduct K-Means clustering based on test documents' representation and report the Purity (\textbf{KM-Purity}) and Normalized Mutual Information (\textbf{KM-NMI}), as in previous works such as \citet{zhao2020neural}. (3) Perplexity: we use document completion perplexity \citep{wallach2009evaluation} to evaluate the predictive ability of topic models. We split each test document into two equal-length folds randomly. Then we compute the Document Completion Perplexity (\textbf{DC-PPL}) on the second fold of documents based on the topic proportion inferred from the first fold, as in previous works such as \citet{dieng-etal-2020-topic}.

\begin{figure*}[!ht]
    \centering
\includegraphics[width=0.85\textwidth,height=\textheight,keepaspectratio]{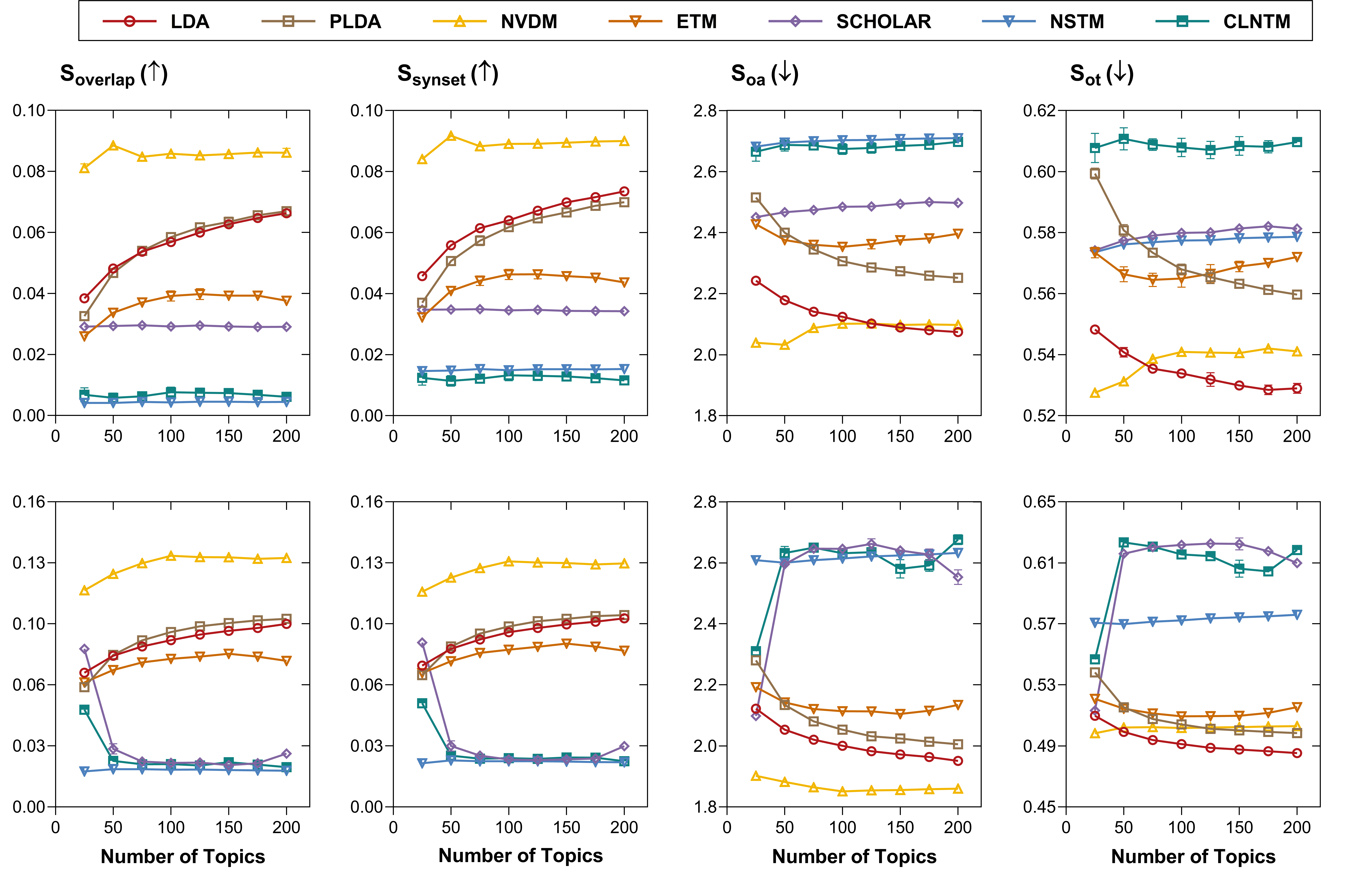}
\vspace{-3mm}
    \caption{Topic models' performance in terms of WALM with \textbf{keywords suggestion} by the LLM on \textbf{20News} (top row) and \textbf{DBpedia} (bottom row). Error bars represent the standard deviation (omitted for values smaller than the symbol size).}
    \label{fig:models_K_noTopic}
\end{figure*}

\begin{figure*}[!ht]
    \centering
\includegraphics[width=0.85\textwidth,height=\textheight,keepaspectratio]{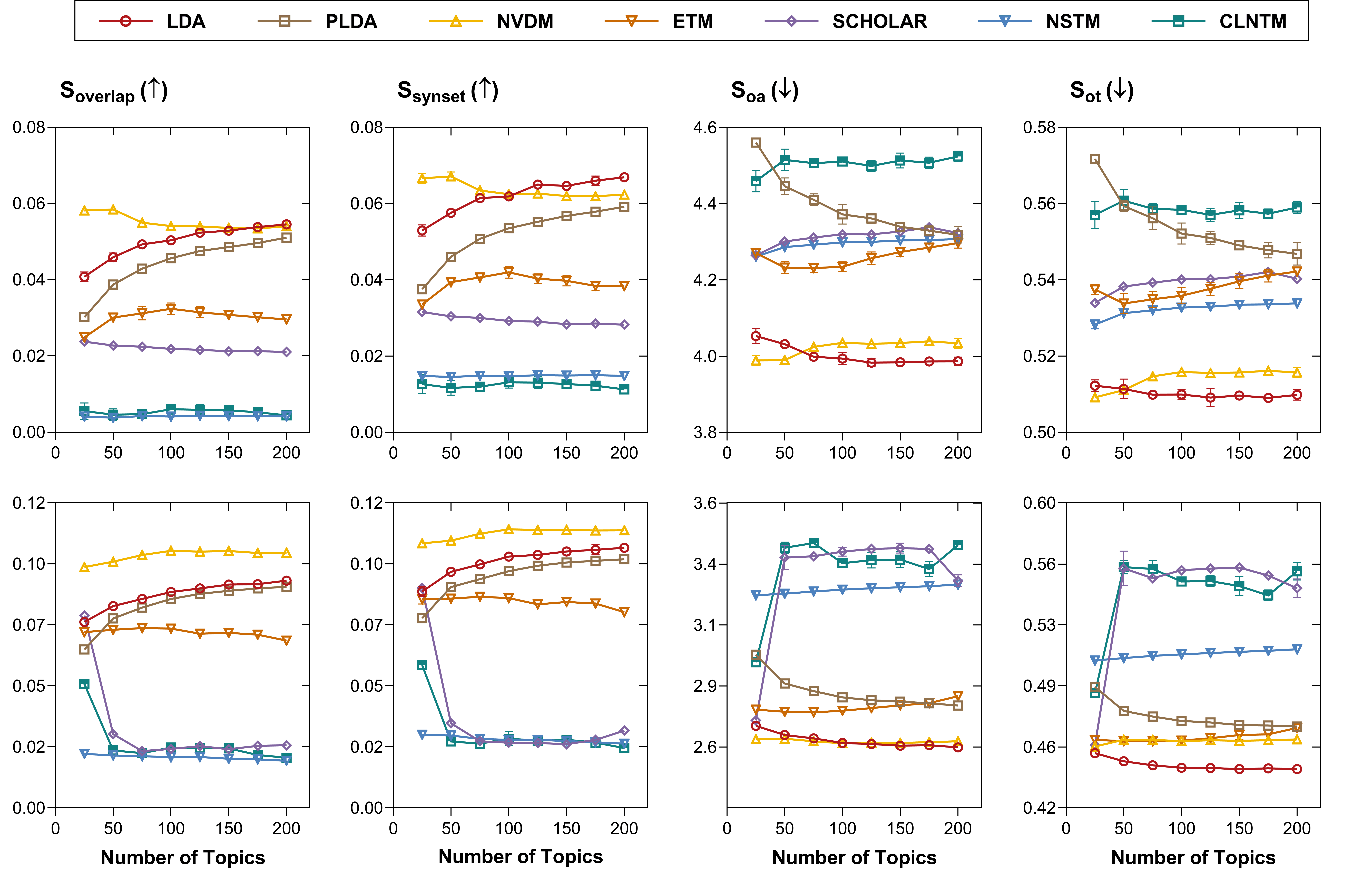}
\vspace{-3mm}
    \caption{Topic models' performance in terms of WALM with \textbf{topic-aware keywords suggestion} by the LLM on \textbf{20News} (top row) and \textbf{DBpedia} (bottom row). Error bars represent the standard deviation (omitted for values smaller than the symbol size).}
    \label{fig:models_K_topic}
\end{figure*}

\begin{figure*}[!t]
    \centering
\includegraphics[width=0.85\textwidth,height=\textheight,keepaspectratio]{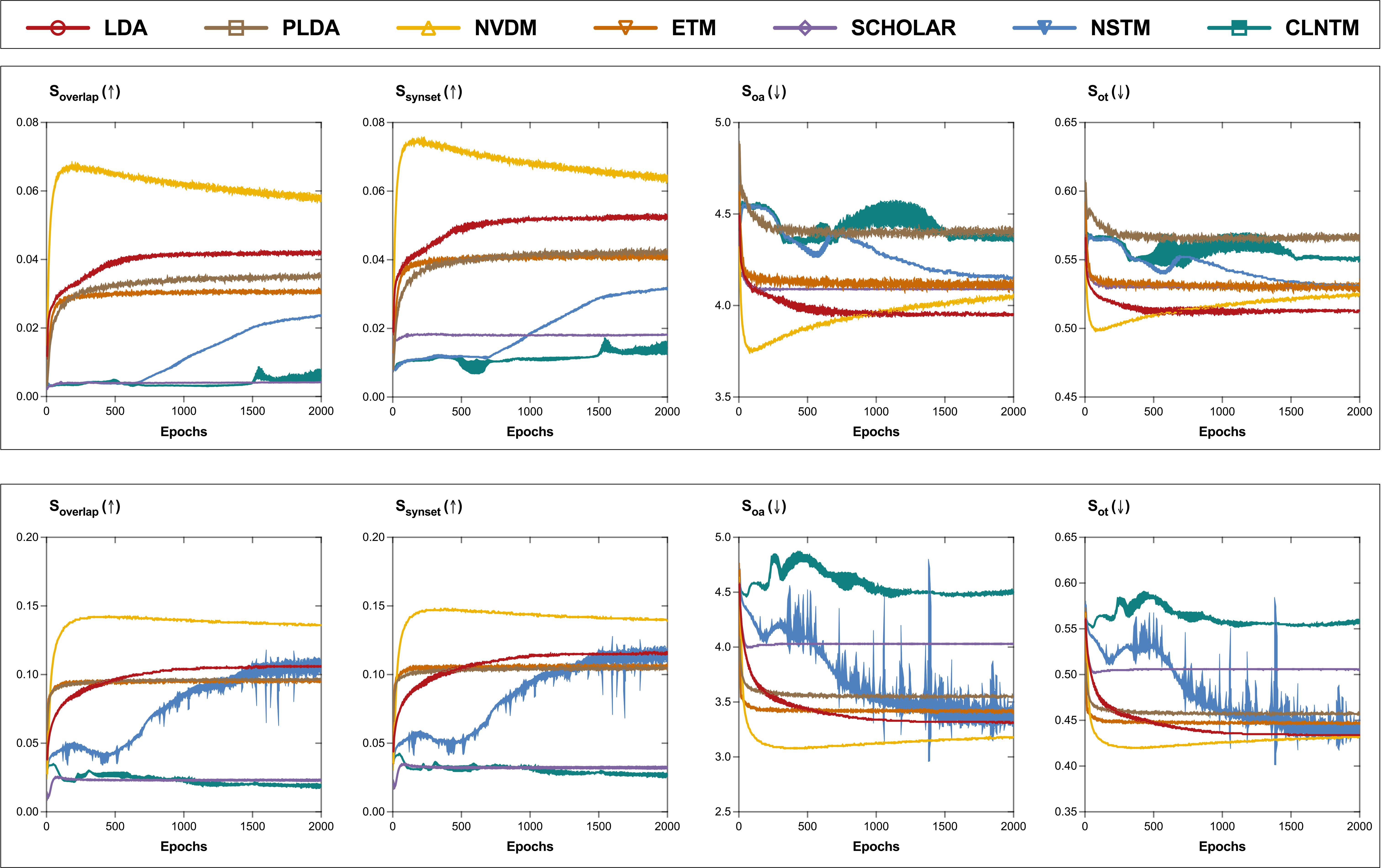}
\vspace{-3mm}
    \caption{Learning curves of topic models in terms of WALM with \textbf{keyword suggestions} (top row) and \textbf{topic-aware keyword suggestions} (bottom row) from the LLM on the 20News test set, with the number of topics set to 50. The area within the error bands represents the standard deviation.}
    \label{fig:learning_curve}
\end{figure*}

\begin{table*}[!ht]
\centering
\resizebox{\textwidth}{!}{
\begin{tabular}{ccl}
\toprule
\textbf{Document} & \textbf{Model} & \textbf{Topical Words} \\
\midrule
\multirow{4}{*}{\pbox{9cm}{\vspace{1mm}It's my understanding that, when you format a magneto-optical disc, (1) the
formatting software installs a driver on the disc, (2) if you insert the
disc in a different drive, then this driver is loaded into the computer's
memory and then controls the drive, and (3) if this driver is incompatible
with the drive, then the disc can not be mounted and/or properly read/written.
Is that correct?}} 
& LDA\_B & drive, disk, card, controller, hard, mb, file, scsi, bios, power \\
& LDA\_C & drive, disk, scsi, hard, card, controller, mb, floppy, ide, sale \\
\cmidrule(l){2-3}
& NVDM\_B & driver, drive, problem, card, time, file, thanks, need, email, work \\
& NVDM\_C & drive, driver, hard, scsi, window, cd, mb, floppy, disc, work \\
\cmidrule(l){2-3} 
& LLM & formatting, magneto-optical, driver, disc, incompatible\\
& LLM (Topic-Aware) & troubleshooting, formatting, incompatibility, magneto-optical, driver, disc, mounting\\
\cmidrule(l){2-3} 
& Human & driver, disc, computer, hardware, software, memory, formatting, incompatible\\
\midrule
\multirow{4}{*}{\pbox{9cm}{\vspace{1cm}Wrong World. Wrong World is a 1985 Australian film directed by Ian Pringle. It was filmed in Nhill and Melbourne in Victoria Australia.}} 
& LDA\_B & film, american, released, directed, football, album, summer, played, team, hospital \\
& LDA\_C & film, played, directed, baseball, league, australian, major, drama, football, award \\
\cmidrule(l){2-3} 
& NVDM\_B & specie, album, school, known, located, north, film, directed, american, released \\
& NVDM\_C & film, album, released, second, south, new, directed, american, australian, known \\
\cmidrule(l){2-3} 
& LLM & world, film, australian, directed, victoria\\
& LLM (Topic-Aware) & film, industry, production, cinema, entertainment\\
\cmidrule(l){2-3} 
& Human & film, movie, directed, director, australian, melbourne, victoria \\
\bottomrule
\end{tabular}
}
\vspace{-3mm}
\caption{Documents' topical words from topic models at the beginning phase (e.g., NVDM\_B, LDA\_B) and convergence phase (e.g., NVDM\_C, LDA\_C) according to WALM, where the number of topics is set to 50.}
\label{qualitive}
\end{table*}

\subsection{Results and Analysis}
\paragraph{Topic Model Evaluation with WALM}
We assess topic models' performance based on our evaluation metrics on both 20News and DBpedia. We have the following observations based on our results illustrated in Figure \ref{fig:models_K_noTopic}: (1) The WALM values of most models on DBpedia show better performance than 20News, which indicates that it is easier for topic models to generate informative topical words for short documents than long documents. (2) The performance ranking indicated by overlap-based metrics (e.g., $S_{\text{overlap}}$ and $S_{\text{synset}}$) and embedding-based metrics (e.g., $S_{\text{oa}}$ and $S_{\text{ot}}$) is slightly different. The reason is that embedding-based metrics consider the semantic distance among words, which can be more flexible than the exact match in overlap-based metrics. (3) It can be observed that there is little improvement from recent NTMs over LDA and NVDM in terms of our joint metrics. The potential reason is that most contemporary NTMs primarily focus on enhancing topic coherence while neglecting the generation of documents, thus showing weak performance in generating topical words of documents as indicated by WALM. (4) When topic-aware keyword suggestion is applied in WALM (Figure \ref{fig:models_K_topic}), the performance ranking of LDA surpasses that of NVDM as the number of topics increases in the long-document dataset (i.e., 20News). This suggests that LDA benefits more from an increased number of topics when generating topic-aware keywords for documents compared to NVDM.

\paragraph{Learning Curves of WALM}
In Figure \ref{fig:learning_curve}, we illustrate the learning curves of topic models in terms of WALM, clearly showing how each metric changes throughout the training process. We observe that most topic models improve with training and eventually converge to a stable state. However, NVDM exhibits overfitting in the later stages of training, as indicated by its WALM scores. Additionally, WALM approaches based on keyword suggestions and topic-aware keyword suggestions exhibit slightly different trends in their learning curves. For instance, LDA surpasses NVDM in the later training stages when topic-aware keywords are used. This suggests that NVDM prioritizes document-level generation while LDA shows stronger awareness of collection-level topics.

\paragraph{Qualitative Analysis on Topical Words for Documents}
We qualitatively investigate the topical words of documents by topic models at different stages in Table \ref{qualitive}, where we randomly sample one document for 20News and DBpedia, respectively. We have the following observations based on our results: (1) The topical words at the beginning phase contain less semantically related words about the documents than those at convergence, which aligns with the learning status (as in Figure \ref{fig:learning_curve}) indicated by WALM. (2) The topical words of NVDM include more words that reveal the documents' main messages than LDA, which aligns with the ranking (as in Figure \ref{fig:models_K_topic}) suggested by WALM. (3) The keywords generated by the LLM are similar to those provided by human annotators for the example documents. (4) By using topic-aware keywords suggestion in WALM, the LLM tends to provide keywords that convey the high-level concepts of the topics. For instance, ``troubleshooting'' is identified for the first example document, and ``entertainment'' for the second, which offers higher-level information from topics besides individual document.

\begin{figure}[!t]
    \centering
    \includegraphics[width=0.48\textwidth]{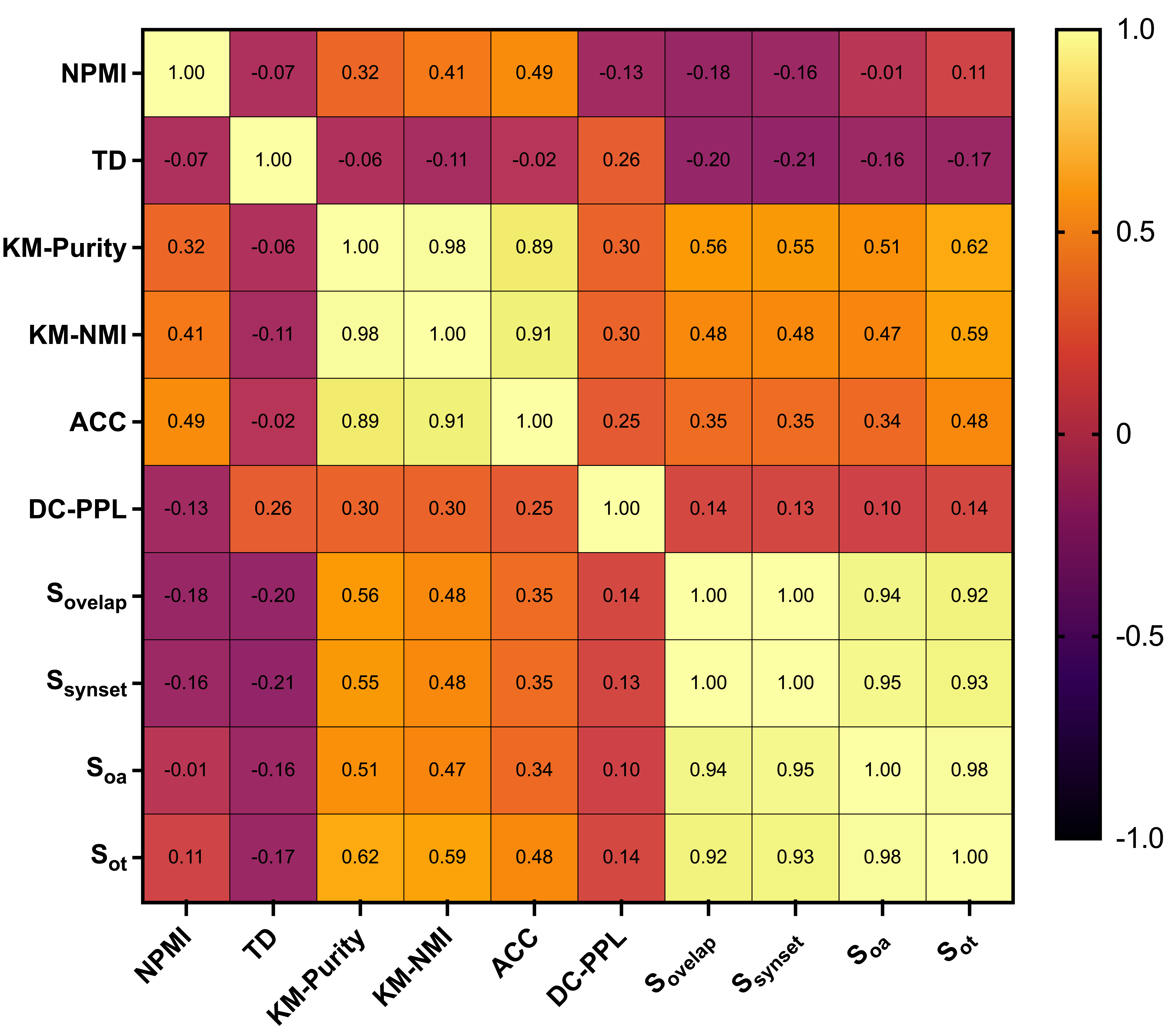}
    \vspace{-6mm}
  \caption{Pearson’s correlation coefficient among evaluation metrics.}
  \label{fig:pear}
\end{figure}

\paragraph{Correlation to Other Metrics} \label{correlation}
We compute Pearson’s correlation coefficients among existing and WALM series metrics, similar to the correlation analysis in previous works such as \citet{doogan2021topic} and \citet{rahimi-etal-2024-contextualized}. Pearson’s correlation coefficients among the metrics are plotted in a heatmap in Figure \ref{fig:pear}. Based on the results, we observe that: (1) WALM variants are highly correlated with each other since they originate from the same mechanism. (2) Compared with perplexity, which also evaluates the entire model based on documents, WALM show weak correlations, suggesting a new family of evaluation metrics. (3) Compared with other types of evaluations, WALM has moderate correlations with document representation metrics (e.g., KM-Purity, KM-NMI, and ACC), and weak correlations with topic quality metrics (e.g., NPMI and TD). This indicates that our joint evaluation metrics take both components into account without relying solely on either one. These observations suggest that WALM can serve as a complementary evaluation method to existing approaches.

\begin{figure}[!t]
    \centering
    \includegraphics[width=0.48\textwidth]{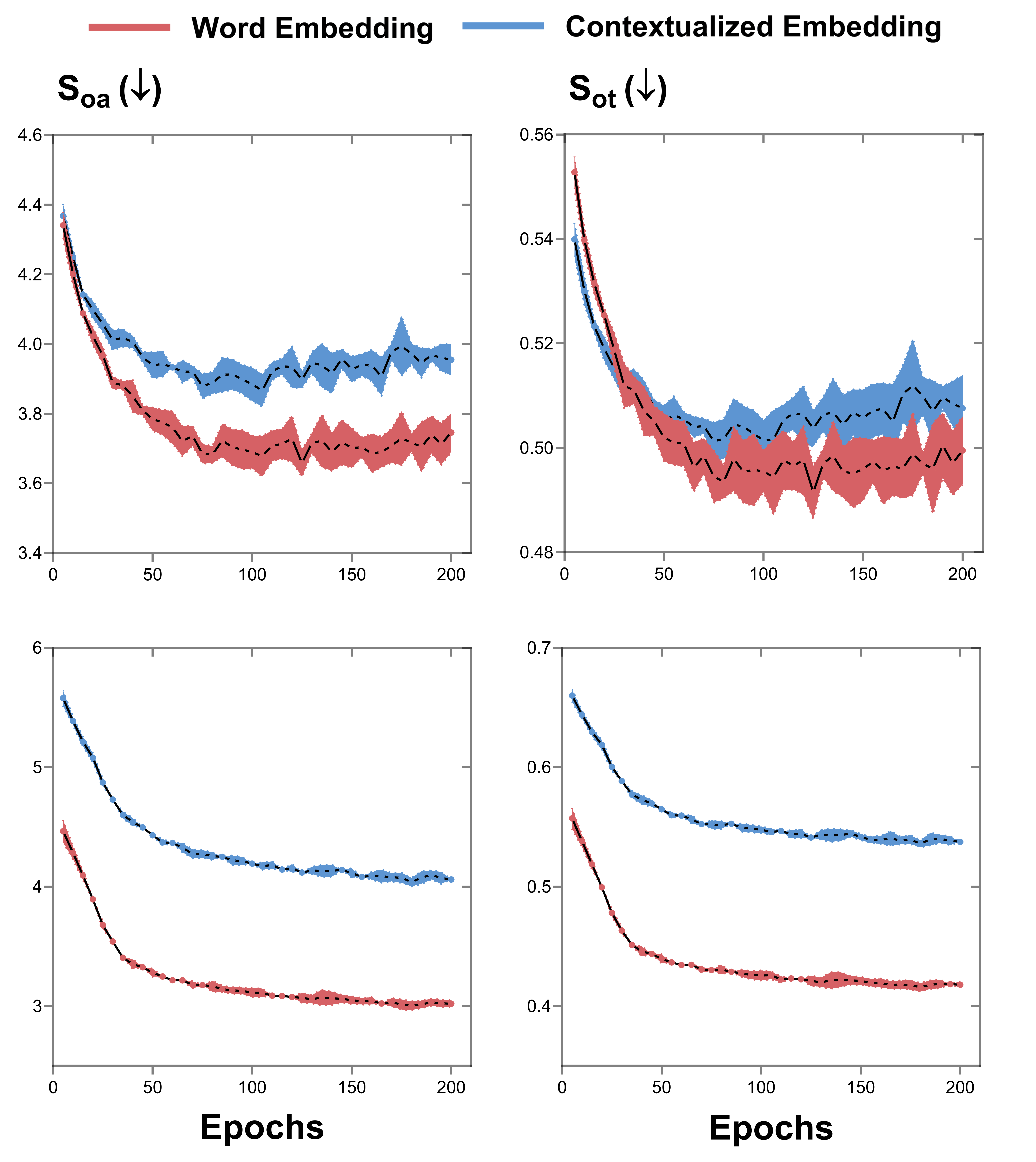}
    \vspace{-6mm}
  \caption{Learning curves of NVDM in terms of embedding-based metrics and their contextualized variants on 20News (top row) and DBpedia (bottom row). The area within the error bands represents the standard deviation.}
  \label{fig:contextualise}
\end{figure}

\begin{figure*}[!t]
    \centering
    \includegraphics[width=0.85\textwidth]{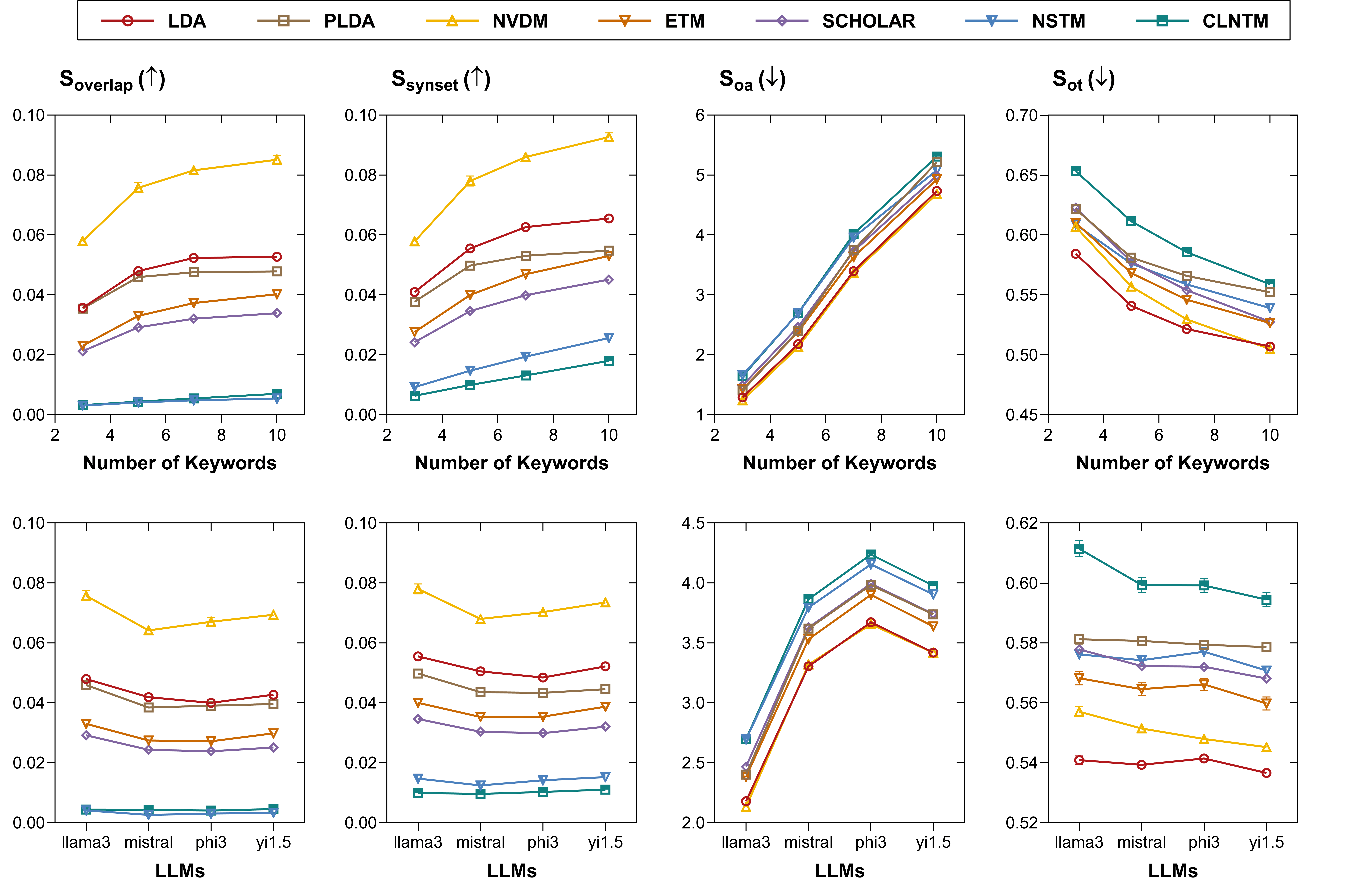}
    \vspace{-3mm}
  \caption{Sensitivity study. \textbf{Top row}: Performance of topic models in terms of WALM with varying numbers of keywords. \textbf{Bottom row}: Performance of topic models in terms of WALM with different LLMs. Experiments are conducted on the 20News dataset with the number of topics set to 50. Error bars represent the standard deviation (omitted for values smaller than the symbol size).}
  \label{hyper}
\end{figure*}

\subsection{Contextualized Embeddings for WALM}\label{c_embedidng}

\paragraph{Obtaining Contextualized Embeddings} 
Recall that in Eq. \ref{oa_score} and Eq. \ref{ot_score}, the cost matrix $\bm{C}$ is constructed using cosine distances between word embeddings. Here, we change our construction of $\bm{C}$ from using static word embeddings from GloVe \citep{pennington2014glove} to the contextualized word embeddings from the LLM, considering that the same word may have different semantic meanings in different contexts. We obtain the contextualized embeddings of a word given a document differently in two cases: (1) When the target word appears in the context document, we take the average embeddings of each occurrence as the contextualized embedding. (2) When there is no occurrence of the target word in the given document, we add an auxiliary sentence to the document in the following format:
\begin{quote}
\itshape
``\textless Given Document\textgreater. This document is talking about \textless Target Word\textgreater.''
\end{quote}
Then, we obtain the contextualized embedding of the target word given the document with the auxiliary sentence. By replacing the global word embeddings with contextualized word embeddings, we have new variants of our embedding-based WALM (i.e., $S_{\text{oa}}$ and $S_{\text{ot}}$), i.e., \textbf{$\bf{S}_{\text{oa\_c}}$} and \textbf{$\bf{S}_{\text{ot\_c}}$}.

\paragraph{Observations}
Since the cost of obtaining contextualized embeddings is high for LLMs, we compute $S_{\text{oa\_c}}$ and $S_{\text{ot\_c}}$ in a case study, where we test NVDM on 100 documents randomly sampled from the test sets of 20News and DBpedia, respectively. We plot the learning curves on test documents in Figure \ref{fig:contextualise}. We observe that using word embeddings or contextualized embeddings in our embedding-based scores exhibits similar trends but with different values on both datasets.

\subsection{Sensitivity Study}
Here, we examine two factors that can influence the WALM scores: the number of keywords generated by the LLM and the choice of the LLM. To investigate the effect of the number of keywords, we vary the number from 3 to 10 and plot the performance ranking of topic models in Figure \ref{hyper} (top row). We observe that, although the values of WALM metrics can vary with different numbers of keywords, the overall performance ranking of the topic models remains largely unaffected by these changes, especially for the overlap-based metrics. To investigate the effect of LLMs, we use different latest LLMs for keyword generation apart from LLAMA3-8B-Instruct, including Mistral-7B-Instruct-v0.3 \citep{jiang2023mistral}, Phi-3-Mini-128K-Instruct \citep{abdin2024phi} and Yi-1.5-9B-Chat \citep{young2024yi}. From the results illustrated in Figure \ref{hyper} (bottom row), we observe that overlap-based metrics show minimal variation with different choices of LLMs, and the performance ranking of the topic models is unaffected in most cases. These observations suggest that the overlap-based metrics are less sensitive to the number of words and the choice of LLMs.


\subsection{Comparisons with Human Annotation}\label{eval_gap}
\paragraph{Evaluation Gap with Human Annotation}
WALM computes the difference between documents' topical words generated by topic models and an LLM, treating the words from the LLM as the ground truth. Here, we investigate the gap between using LLM and human judgment as the true topical words in WALM. To quantify this gap, we use the following calculation:
\begin{equation}\label{eq:eval_gap} \tag{12}
G:=\frac{|S(\text{LLM as Truth})-S(\text{Human as Truth})|}{S(\text{Human as Truth})},
\end{equation}
where $S$ is the WLAM scores we propose in Sec. \ref{sfunction} (and the contextualized variants in Sec. \ref{c_embedidng}). Intuitively, the gap function measures the difference between using ground truth (e.g., keywords) from the LLM and those from human annotators. We empirically observe that topical words from topic models consistently differ from those identified by humans, so the denominator in Eq. \ref{eq:eval_gap} will not be zero.

As human annotation is expensive for large-scale investigation, we randomly sample 200 test documents from 20News and DBpedia as a case study. We engaged three English speakers as annotators, trained with a few examples, to provide keywords that capture the main points of each document. Then, given a trained topic model, we compute the gap between using the words from the LLM and human in our metrics using Eq. \ref{eq:eval_gap}.

\begin{figure}[!t]
    \centering
    \includegraphics[width=0.5\textwidth]{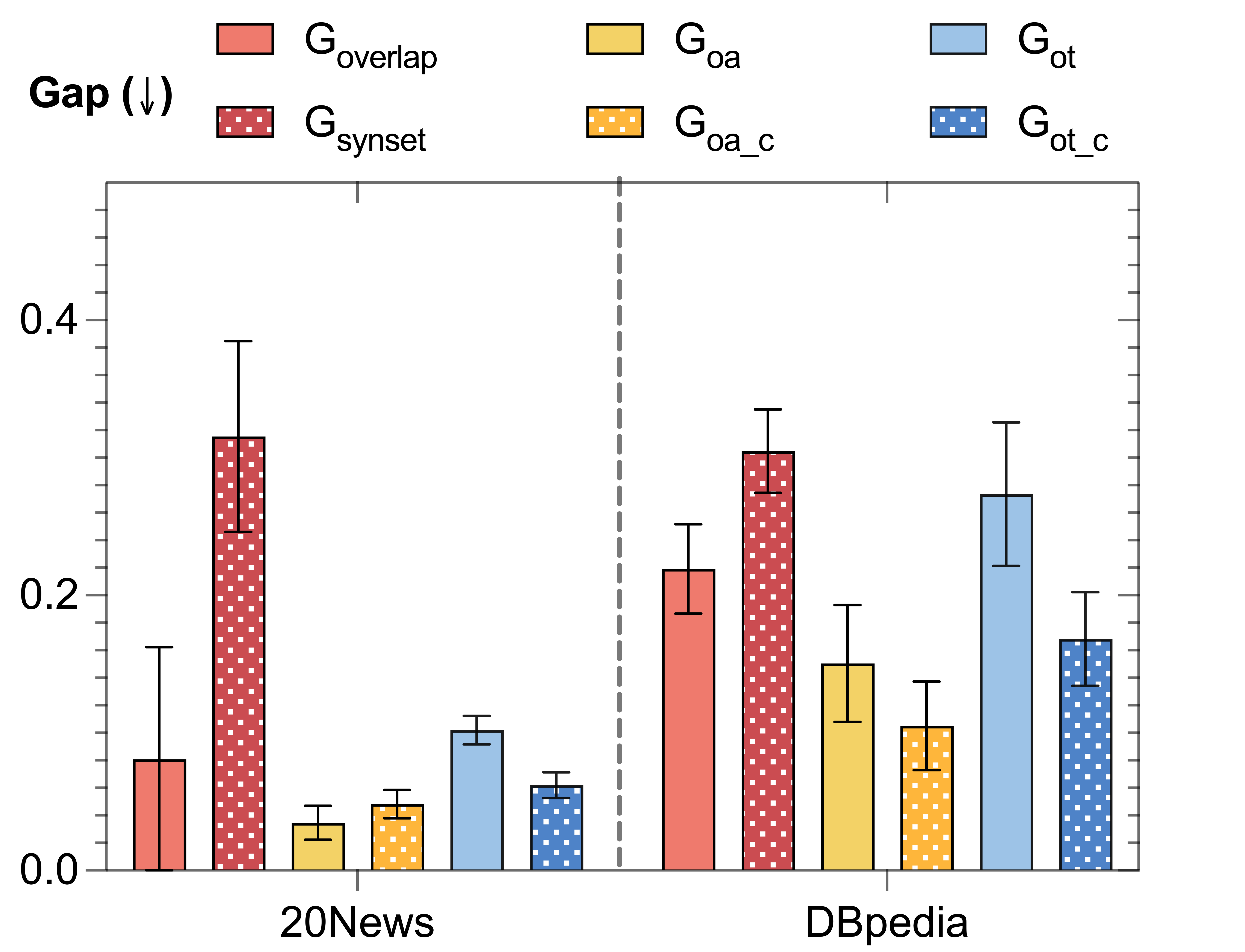}
    \vspace{-6mm}
  \caption{Evaluation gap between using the LLM and human judgment as the ``true'' topical words. Error bars represent the standard deviation.}
  \label{human}
\end{figure}

The results are illustrated in Figure \ref{human}, where the evaluated model is NVDM with $K=50$ trained on 20News and DBpedia, respectively. We have the following observations based on the results: (1) Comparing the datasets, the gap between using human judgment and the LLM in 20News is lower than in DBpedia in most cases. This indicates that for long documents such as those in 20News, the topical words generated by the LLM are closer to human judgment than in short documents in DBpedia. (2) Comparing the metrics, $S_{\text{oa}}$ exhibits the lowest gap among WALM metrics, with a gap value of 0.03 and 0.15 on 20News and DBpedia, respectively. This shows the effectiveness of using the LLM as a proxy for human judgment when applied in $S_{\text{oa}}$. (3) Comparing the embeddings, using contextualized embeddings from the LLM can further narrow the evaluation gap for $S_{\text{oa}}$ and $S_{\text{ot}}$ on short documents.

\begin{table}[!t]
\centering
\resizebox{0.48\textwidth}{!}{
\begin{tabular}{ccccc}
\toprule
& $S_{\text{overlap}}$ & $S_{\text{synset}}$ & $S_{\text{oa}}$ & $S_{\text{ot}}$ \\
\midrule
5-word suggestion & 0.55 & 0.50 & 0.57 & 0.63 \\
10-word suggestion & 0.52 & 0.58 & 0.56 & 0.68 \\
\bottomrule
\end{tabular}
}
\vspace{-3mm}
\caption{Pearson's correlation coefficient between WALM using LLM-generated keywords and human annotations as the ground truth on the 500N-KPCrowd dataset.}
\label{huamn_cor}
\end{table}

\paragraph{Correlation with Human Annotation} 
We use an existing annotated dataset, 500N-KPCrowd \citep{MARUJO12.672} for the keyphrase extraction task \citep{hasan2014automatic}, where each test document is paired with labeled keywords. We run LDA on the training documents and infer the topical words for the test documents, then compute the Pearson's correlation coefficient between the WALM scores using the LLM-generated keywords and the test labels as the ground truth. The results are illustrated in Table. \ref{huamn_cor}. We observe that (1) using keywords from the LLM in WALM scores correlates with using the labeled keyphrases, and (2) the correlation can potentially improve when more keywords are included in the LLM's suggestions.

\section{Conclusion}
In this work, we propose WALM for topic model evaluation, which takes both topic and document representation quality into account jointly. WALM measures the agreement between the topical words generated by topic models and those from the LLM for given documents. The topical words from the LLM are obtained through keyword prompting or topic-aware keyword prompting, with the latter tending to capture higher-level information. To quantify the agreement between word sets, we propose different calculations, including overlap-based and embedding-based metrics. Our experiments demonstrate that the WALM series effectively reflect the capability of topic models to provide semantic summaries of documents. We show that WALM metrics align with human judgment and can serve as an informative complementary method for topic model evaluation. We suggest that overlap-based metrics demonstrate better sensitivity handling, while embedding-based metrics show a smaller evaluation gap. A potential risk of using WALM is that models chasing this metric only may be affected by the bias of LLMs. To mitigate the risk, we suggest using WALM with other metrics together.


\section*{Acknowledgments}
We thank the anonymous reviewers and the action editor, Michael Elhadad, for their valuable feedback, which has significantly strengthened this work.

\bibliography{main}

\begin{thebibliography}{72}
\expandafter\ifx\csname natexlab\endcsname\relax\def\natexlab#1{#1}\fi

\bibitem[{Abdin et~al.(2024)Abdin, Jacobs, Awan, Aneja, Awadallah, Awadalla, Bach, Bahree, Bakhtiari, Behl et~al.}]{abdin2024phi}
Marah Abdin, Sam~Ade Jacobs, Ammar~Ahmad Awan, Jyoti Aneja, Ahmed Awadallah, Hany Awadalla, Nguyen Bach, Amit Bahree, Arash Bakhtiari, Harkirat Behl, et~al. 2024.
\newblock Phi-3 technical report: A highly capable language model locally on your phone.
\newblock \emph{arXiv preprint arXiv:2404.14219}.

\bibitem[{Auer et~al.(2007)Auer, Bizer, Kobilarov, Lehmann, Cyganiak, and Ives}]{auer2007dbpedia}
S{\"o}ren Auer, Christian Bizer, Georgi Kobilarov, Jens Lehmann, Richard Cyganiak, and Zachary Ives. 2007.
\newblock Dbpedia: A nucleus for a web of open data.
\newblock In \emph{international semantic web conference}, pages 722--735. Springer.

\bibitem[{Bai et~al.(2024)Bai, Wu, Stojkovic, and Tsioutsiouliklis}]{10.1145/3627673.3680093}
Xiao Bai, Xue Wu, Ivan Stojkovic, and Kostas Tsioutsiouliklis. 2024.
\newblock Leveraging large language models for improving keyphrase generation for contextual targeting.
\newblock In \emph{Proceedings of the 33rd ACM International Conference on Information and Knowledge Management}, CIKM '24, page 4349–4357, New York, NY, USA. Association for Computing Machinery.

\bibitem[{Bhatia et~al.(2017)Bhatia, Lau, and Baldwin}]{bhatia-etal-2017-automatic}
Shraey Bhatia, Jey~Han Lau, and Timothy Baldwin. 2017.
\newblock An automatic approach for document-level topic model evaluation.
\newblock In \emph{Proceedings of the 21st Conference on Computational Natural Language Learning ({C}o{NLL} 2017)}, pages 206--215, Vancouver, Canada. Association for Computational Linguistics.

\bibitem[{Bhatia et~al.(2018)Bhatia, Lau, and Baldwin}]{bhatia-etal-2018-topic}
Shraey Bhatia, Jey~Han Lau, and Timothy Baldwin. 2018.
\newblock Topic intrusion for automatic topic model evaluation.
\newblock In \emph{Proceedings of the 2018 Conference on Empirical Methods in Natural Language Processing}, pages 844--849, Brussels, Belgium. Association for Computational Linguistics.

\bibitem[{Blei et~al.(2010)Blei, Griffiths, and Jordan}]{10.1145/1667053.1667056}
David~M. Blei, Thomas~L. Griffiths, and Michael~I. Jordan. 2010.
\newblock The nested chinese restaurant process and bayesian nonparametric inference of topic hierarchies.
\newblock \emph{J. ACM}, 57(2).

\bibitem[{Blei et~al.(2003)Blei, Ng, and Jordan}]{blei2003latent}
David~M Blei, Andrew~Y Ng, and Michael~I Jordan. 2003.
\newblock Latent dirichlet allocation.
\newblock \emph{Journal of machine Learning research}, 3(Jan):993--1022.

\bibitem[{Brown et~al.(2020)Brown, Mann, Ryder, Subbiah, Kaplan, Dhariwal, Neelakantan, Shyam, Sastry, Askell et~al.}]{brown2020language}
Tom Brown, Benjamin Mann, Nick Ryder, Melanie Subbiah, Jared~D Kaplan, Prafulla Dhariwal, Arvind Neelakantan, Pranav Shyam, Girish Sastry, Amanda Askell, et~al. 2020.
\newblock Language models are few-shot learners.
\newblock \emph{Advances in neural information processing systems}, 33:1877--1901.

\bibitem[{Bui et~al.(2022)Bui, Le, Tran, Zhao, and Phung}]{buiunified}
Anh~Tuan Bui, Trung Le, Quan~Hung Tran, He~Zhao, and Dinh Phung. 2022.
\newblock A unified {W}asserstein distributional robustness framework for adversarial training.
\newblock In \emph{International Conference on Learning Representations}.

\bibitem[{Buntine(2009)}]{buntine2009estimating}
Wray Buntine. 2009.
\newblock Estimating likelihoods for topic models.
\newblock In \emph{Asian conference on machine learning}, pages 51--64. Springer.

\bibitem[{Card et~al.(2018)Card, Tan, and Smith}]{card-etal-2018-neural}
Dallas Card, Chenhao Tan, and Noah~A. Smith. 2018.
\newblock Neural models for documents with metadata.
\newblock In \emph{Proceedings of the 56th Annual Meeting of the Association for Computational Linguistics (Volume 1: Long Papers)}, pages 2031--2040, Melbourne, Australia. Association for Computational Linguistics.

\bibitem[{Chang et~al.(2009)Chang, Gerrish, Wang, Boyd-Graber, and Blei}]{chang2009reading}
Jonathan Chang, Sean Gerrish, Chong Wang, Jordan Boyd-Graber, and David Blei. 2009.
\newblock Reading tea leaves: How humans interpret topic models.
\newblock \emph{Advances in neural information processing systems}, 22.

\bibitem[{Chowdhery et~al.(2024)Chowdhery, Narang, Devlin, Bosma, Mishra, Roberts, Barham, Chung, Sutton, Gehrmann, Schuh, Shi, Tsvyashchenko, Maynez, Rao, Barnes, Tay, Shazeer, Prabhakaran, Reif, Du, Hutchinson, Pope, Bradbury, Austin, Isard, Gur-Ari, Yin, Duke, Levskaya, Ghemawat, Dev, Michalewski, Garcia, Misra, Robinson, Fedus, Zhou, Ippolito, Luan, Lim, Zoph, Spiridonov, Sepassi, Dohan, Agrawal, Omernick, Dai, Pillai, Pellat, Lewkowycz, Moreira, Child, Polozov, Lee, Zhou, Wang, Saeta, Diaz, Firat, Catasta, Wei, Meier-Hellstern, Eck, Dean, Petrov, and Fiedel}]{10.5555/3648699.3648939}
Aakanksha Chowdhery, Sharan Narang, Jacob Devlin, Maarten Bosma, Gaurav Mishra, Adam Roberts, Paul Barham, Hyung~Won Chung, Charles Sutton, Sebastian Gehrmann, Parker Schuh, Kensen Shi, Sashank Tsvyashchenko, Joshua Maynez, Abhishek Rao, Parker Barnes, Yi~Tay, Noam Shazeer, Vinodkumar Prabhakaran, Emily Reif, Nan Du, Ben Hutchinson, Reiner Pope, James Bradbury, Jacob Austin, Michael Isard, Guy Gur-Ari, Pengcheng Yin, Toju Duke, Anselm Levskaya, Sanjay Ghemawat, Sunipa Dev, Henryk Michalewski, Xavier Garcia, Vedant Misra, Kevin Robinson, Liam Fedus, Denny Zhou, Daphne Ippolito, David Luan, Hyeontaek Lim, Barret Zoph, Alexander Spiridonov, Ryan Sepassi, David Dohan, Shivani Agrawal, Mark Omernick, Andrew~M. Dai, Thanumalayan~Sankaranarayana Pillai, Marie Pellat, Aitor Lewkowycz, Erica Moreira, Rewon Child, Oleksandr Polozov, Katherine Lee, Zongwei Zhou, Xuezhi Wang, Brennan Saeta, Mark Diaz, Orhan Firat, Michele Catasta, Jason Wei, Kathy Meier-Hellstern, Douglas Eck, Jeff Dean, Slav Petrov, and Noah Fiedel.
  2024.
\newblock Palm: scaling language modeling with pathways.
\newblock \emph{J. Mach. Learn. Res.}, 24(1).

\bibitem[{Dieng et~al.(2020)Dieng, Ruiz, and Blei}]{dieng-etal-2020-topic}
Adji~B. Dieng, Francisco J.~R. Ruiz, and David~M. Blei. 2020.
\newblock Topic modeling in embedding spaces.
\newblock \emph{Transactions of the Association for Computational Linguistics}, 8:439--453.

\bibitem[{Doogan and Buntine(2021)}]{doogan2021topic}
Caitlin Doogan and Wray Buntine. 2021.
\newblock Topic model or topic twaddle? re-evaluating semantic interpretability measures.
\newblock In \emph{Proceedings of the 2021 conference of the North American chapter of the association for computational linguistics: human language technologies}, pages 3824--3848.

\bibitem[{Flamary et~al.(2021)Flamary, Courty, Gramfort, Alaya, Boisbunon, Chambon, Chapel, Corenflos, Fatras, Fournier et~al.}]{flamary2021pot}
R{\'e}mi Flamary, Nicolas Courty, Alexandre Gramfort, Mokhtar~Z Alaya, Aur{\'e}lie Boisbunon, Stanislas Chambon, Laetitia Chapel, Adrien Corenflos, Kilian Fatras, Nemo Fournier, et~al. 2021.
\newblock Pot: Python optimal transport.
\newblock \emph{Journal of Machine Learning Research}, 22(78):1--8.

\bibitem[{Gan et~al.(2015)Gan, Henao, Carlson, and Carin}]{gan2015learning}
Zhe Gan, R.~Henao, D.~Carlson, and Lawrence Carin. 2015.
\newblock Learning deep sigmoid belief networks with data augmentation.
\newblock In \emph{AISTATS}, pages 268--276.

\bibitem[{Gao et~al.(2024)Gao, Zhao, dan Guo, and Zha}]{gao2024distribution}
Jintong Gao, He~Zhao, Dan dan Guo, and Hongyuan Zha. 2024.
\newblock Distribution alignment optimization through neural collapse for long-tailed classification.
\newblock In \emph{Forty-first International Conference on Machine Learning}.

\bibitem[{Ge et~al.(2021)Ge, Liu, Li, Yoshie, and Sun}]{ge2021ota}
Zheng Ge, Songtao Liu, Zeming Li, Osamu Yoshie, and Jian Sun. 2021.
\newblock {OTA}: Optimal transport assignment for object detection.
\newblock In \emph{CVPR}, pages 303--312.

\bibitem[{Grootendorst(2022)}]{grootendorst2022bertopic}
Maarten Grootendorst. 2022.
\newblock Bertopic: Neural topic modeling with a class-based tf-idf procedure.
\newblock \emph{arXiv preprint arXiv:2203.05794}.

\bibitem[{Guo et~al.(2022)Guo, Tian, Zhang, Zhou, and Zha}]{danlearning2022}
Dandan Guo, Long Tian, Minghe Zhang, Mingyuan Zhou, and Hongyuan Zha. 2022.
\newblock Learning prototype-oriented set representations for meta-learning.
\newblock In \emph{International Conference on Learning Representations}.

\bibitem[{Hasan and Ng(2014)}]{hasan2014automatic}
Kazi~Saidul Hasan and Vincent Ng. 2014.
\newblock Automatic keyphrase extraction: A survey of the state of the art.
\newblock In \emph{Proceedings of the 52nd Annual Meeting of the Association for Computational Linguistics (Volume 1: Long Papers)}, pages 1262--1273.

\bibitem[{Hoover et~al.(2021)Hoover, Du, Sordoni, and O{'}Donnell}]{hoover-etal-2021-linguistic}
Jacob~Louis Hoover, Wenyu Du, Alessandro Sordoni, and Timothy~J. O{'}Donnell. 2021.
\newblock Linguistic dependencies and statistical dependence.
\newblock In \emph{Proceedings of the 2021 Conference on Empirical Methods in Natural Language Processing}, pages 2941--2963, Online and Punta Cana, Dominican Republic. Association for Computational Linguistics.

\bibitem[{Jiang et~al.(2023)Jiang, Sablayrolles, Mensch, Bamford, Chaplot, Casas, Bressand, Lengyel, Lample, Saulnier et~al.}]{jiang2023mistral}
Albert~Q Jiang, Alexandre Sablayrolles, Arthur Mensch, Chris Bamford, Devendra~Singh Chaplot, Diego de~las Casas, Florian Bressand, Gianna Lengyel, Guillaume Lample, Lucile Saulnier, et~al. 2023.
\newblock Mistral 7b.
\newblock \emph{arXiv preprint arXiv:2310.06825}.

\bibitem[{Kuhn(1955)}]{kuhn1955hungarian}
Harold~W Kuhn. 1955.
\newblock The hungarian method for the assignment problem.
\newblock \emph{Naval research logistics quarterly}, 2(1-2):83--97.

\bibitem[{Lang(1995)}]{lang1995newsweeder}
Ken Lang. 1995.
\newblock Newsweeder: Learning to filter netnews.
\newblock In \emph{Machine learning proceedings 1995}, pages 331--339. Elsevier.

\bibitem[{Larochelle and Lauly(2012)}]{larochelle2012neural}
Hugo Larochelle and Stanislas Lauly. 2012.
\newblock A neural autoregressive topic model.
\newblock \emph{Advances in Neural Information Processing Systems}, 25.

\bibitem[{Lau et~al.(2014)Lau, Newman, and Baldwin}]{lau2014machine}
Jey~Han Lau, David Newman, and Timothy Baldwin. 2014.
\newblock Machine reading tea leaves: Automatically evaluating topic coherence and topic model quality.
\newblock In \emph{Proceedings of the 14th Conference of the European Chapter of the Association for Computational Linguistics}, pages 530--539.

\bibitem[{Laureate et~al.(2023)Laureate, Buntine, and Linger}]{laureate2023systematic}
Caitlin Doogan~Poet Laureate, Wray Buntine, and Henry Linger. 2023.
\newblock A systematic review of the use of topic models for short text social media analysis.
\newblock \emph{Artificial Intelligence Review}, pages 1--33.

\bibitem[{Liu et~al.(2016)Liu, Tang, Dong, Yao, and Zhou}]{liu2016overview}
Lin Liu, Lin Tang, Wen Dong, Shaowen Yao, and Wei Zhou. 2016.
\newblock An overview of topic modeling and its current applications in bioinformatics.
\newblock \emph{SpringerPlus}, 5(1):1--22.

\bibitem[{Lucas et~al.(2019)Lucas, Tucker, Grosse, and Norouzi}]{lucas2019don}
James Lucas, George Tucker, Roger~B Grosse, and Mohammad Norouzi. 2019.
\newblock Don't blame the elbo! a linear vae perspective on posterior collapse.
\newblock \emph{Advances in Neural Information Processing Systems}, 32.

\bibitem[{Maragheh et~al.(2023)Maragheh, Fang, Irugu, Parikh, Cho, Xu, Sukumar, Patel, Korpeoglu, Kumar et~al.}]{maragheh2023llm}
Reza~Yousefi Maragheh, Chenhao Fang, Charan~Chand Irugu, Parth Parikh, Jason Cho, Jianpeng Xu, Saranyan Sukumar, Malay Patel, Evren Korpeoglu, Sushant Kumar, et~al. 2023.
\newblock Llm-take: theme-aware keyword extraction using large language models.
\newblock In \emph{2023 IEEE International Conference on Big Data (BigData)}, pages 4318--4324. IEEE.

\bibitem[{Marujo et~al.(2012)Marujo, Gershman, Carbonell, Frederking, and Neto}]{MARUJO12.672}
Luís Marujo, Anatole Gershman, Jaime Carbonell, Robert Frederking, and JoaÌƒo~P. Neto. 2012.
\newblock Supervised topical key phrase extraction of news stories using crowdsourcing, light filtering and co-reference normalization.
\newblock In \emph{Proceedings of the Eight International Conference on Language Resources and Evaluation (LREC'12)}, Istanbul, Turkey. European Language Resources Association (ELRA).

\bibitem[{Miao et~al.(2017)Miao, Grefenstette, and Blunsom}]{miao2017discovering}
Yishu Miao, Edward Grefenstette, and Phil Blunsom. 2017.
\newblock Discovering discrete latent topics with neural variational inference.
\newblock In \emph{International conference on machine learning}, pages 2410--2419. PMLR.

\bibitem[{Mikolov et~al.(2013)Mikolov, Chen, Corrado, and Dean}]{mikolov2013efficient}
Tomas Mikolov, Kai Chen, Greg Corrado, and Jeffrey Dean. 2013.
\newblock Efficient estimation of word representations in vector space.
\newblock \emph{arXiv preprint arXiv:1301.3781}.

\bibitem[{Miller(1995)}]{miller1995wordnet}
George~A Miller. 1995.
\newblock Wordnet: a lexical database for english.
\newblock \emph{Communications of the ACM}, 38(11):39--41.

\bibitem[{Mimno et~al.(2011)Mimno, Wallach, Talley, Leenders, and McCallum}]{mimno2011optimizing}
David Mimno, Hanna Wallach, Edmund Talley, Miriam Leenders, and Andrew McCallum. 2011.
\newblock Optimizing semantic coherence in topic models.
\newblock In \emph{Proceedings of the 2011 conference on empirical methods in natural language processing}, pages 262--272.

\bibitem[{Newman et~al.(2010)Newman, Lau, Grieser, and Baldwin}]{newman2010automatic}
David Newman, Jey~Han Lau, Karl Grieser, and Timothy Baldwin. 2010.
\newblock Automatic evaluation of topic coherence.
\newblock In \emph{Human language technologies: The 2010 annual conference of the North American chapter of the association for computational linguistics}, pages 100--108.

\bibitem[{Nguyen and Luu(2021)}]{nguyen2021contrastive}
Thong Nguyen and Anh~Tuan Luu. 2021.
\newblock Contrastive learning for neural topic model.
\newblock \emph{Advances in neural information processing systems}, 34:11974--11986.

\bibitem[{Nguyen et~al.(2021)Nguyen, Le, Zhao, Tran, Nguyen, and Phung}]{nguyen2021most}
Tuan Nguyen, Trung Le, He~Zhao, Quan~Hung Tran, Truyen Nguyen, and Dinh Phung. 2021.
\newblock Most: Multi-source domain adaptation via optimal transport for student-teacher learning.
\newblock In \emph{UAI}, pages 225--235.

\bibitem[{Nikolenko(2016)}]{nikolenko2016topic}
Sergey~I Nikolenko. 2016.
\newblock Topic quality metrics based on distributed word representations.
\newblock In \emph{Proceedings of the 39th International ACM SIGIR conference on Research and Development in Information Retrieval}, pages 1029--1032.

\bibitem[{Paisley et~al.(2015)Paisley, Wang, Blei, and Jordan}]{paisley2015nested}
John Paisley, Chong Wang, David~M Blei, and Michael~I Jordan. 2015.
\newblock Nested hierarchical {D}irichlet processes.
\newblock \emph{TPAMI}, 37(2):256--270.

\bibitem[{Pennington et~al.(2014)Pennington, Socher, and Manning}]{pennington2014glove}
Jeffrey Pennington, Richard Socher, and Christopher~D Manning. 2014.
\newblock Glove: Global vectors for word representation.
\newblock In \emph{Proceedings of the 2014 conference on empirical methods in natural language processing (EMNLP)}, pages 1532--1543.

\bibitem[{Pham et~al.(2024)Pham, Hoyle, Sun, Resnik, and Iyyer}]{pham-etal-2024-topicgpt}
Chau Pham, Alexander Hoyle, Simeng Sun, Philip Resnik, and Mohit Iyyer. 2024.
\newblock {T}opic{GPT}: A prompt-based topic modeling framework.
\newblock In \emph{Proceedings of the 2024 Conference of the North American Chapter of the Association for Computational Linguistics: Human Language Technologies (Volume 1: Long Papers)}, pages 2956--2984, Mexico City, Mexico. Association for Computational Linguistics.

\bibitem[{Rahimi et~al.(2024)Rahimi, Mimno, Hoover, Naacke, Constantin, and Amann}]{rahimi-etal-2024-contextualized}
Hamed Rahimi, David Mimno, Jacob Hoover, Hubert Naacke, Camelia Constantin, and Bernd Amann. 2024.
\newblock Contextualized topic coherence metrics.
\newblock In \emph{Findings of the Association for Computational Linguistics: EACL 2024}, pages 1760--1773, St. Julian{'}s, Malta. Association for Computational Linguistics.

\bibitem[{Reisenbichler and Reutterer(2019)}]{reisenbichler2019topic}
Martin Reisenbichler and Thomas Reutterer. 2019.
\newblock Topic modeling in marketing: recent advances and research opportunities.
\newblock \emph{Journal of Business Economics}, 89(3):327--356.

\bibitem[{R{\"o}der et~al.(2015)R{\"o}der, Both, and Hinneburg}]{roder2015exploring}
Michael R{\"o}der, Andreas Both, and Alexander Hinneburg. 2015.
\newblock Exploring the space of topic coherence measures.
\newblock In \emph{Proceedings of the eighth ACM international conference on Web search and data mining}, pages 399--408.

\bibitem[{Sia et~al.(2020)Sia, Dalmia, and Mielke}]{sia-etal-2020-tired}
Suzanna Sia, Ayush Dalmia, and Sabrina~J. Mielke. 2020.
\newblock Tired of topic models? clusters of pretrained word embeddings make for fast and good topics too!
\newblock In \emph{Proceedings of the 2020 Conference on Empirical Methods in Natural Language Processing (EMNLP)}, pages 1728--1736, Online. Association for Computational Linguistics.

\bibitem[{Song et~al.(2023)Song, Geng, Yao, Lu, Feng, and Jing}]{song2023large}
Mingyang Song, Xuelian Geng, Songfang Yao, Shilong Lu, Yi~Feng, and Liping Jing. 2023.
\newblock Large language models as zero-shot keyphrase extractor: A preliminary empirical study.
\newblock \emph{arXiv preprint arXiv:2312.15156}.

\bibitem[{Srivastava and Sutton(2017)}]{srivastava2017autoencoding}
Akash Srivastava and Charles Sutton. 2017.
\newblock Autoencoding variational inference for topic models.
\newblock In \emph{5th International Conference on Learning Representations}.

\bibitem[{Stammbach et~al.(2023)Stammbach, Zouhar, Hoyle, Sachan, and Ash}]{stammbach-etal-2023-revisiting}
Dominik Stammbach, Vil{\'e}m Zouhar, Alexander Hoyle, Mrinmaya Sachan, and Elliott Ash. 2023.
\newblock Revisiting automated topic model evaluation with large language models.
\newblock In \emph{Proceedings of the 2023 Conference on Empirical Methods in Natural Language Processing}, pages 9348--9357, Singapore. Association for Computational Linguistics.

\bibitem[{Tang et~al.(2023)Tang, Sun, Idnay, Nestor, Soroush, Elias, Xu, Ding, Durrett, Rousseau et~al.}]{tang2023evaluating}
Liyan Tang, Zhaoyi Sun, Betina Idnay, Jordan~G Nestor, Ali Soroush, Pierre~A Elias, Ziyang Xu, Ying Ding, Greg Durrett, Justin~F Rousseau, et~al. 2023.
\newblock Evaluating large language models on medical evidence summarization.
\newblock \emph{npj Digital Medicine}, 6(1):158.

\bibitem[{Thoppilan et~al.(2022)Thoppilan, De~Freitas, Hall, Shazeer, Kulshreshtha, Cheng, Jin, Bos, Baker, Du et~al.}]{thoppilan2022lamda}
Romal Thoppilan, Daniel De~Freitas, Jamie Hall, Noam Shazeer, Apoorv Kulshreshtha, Heng-Tze Cheng, Alicia Jin, Taylor Bos, Leslie Baker, Yu~Du, et~al. 2022.
\newblock Lamda: Language models for dialog applications.
\newblock \emph{arXiv preprint arXiv:2201.08239}.

\bibitem[{Touvron et~al.(2023{\natexlab{a}})Touvron, Lavril, Izacard, Martinet, Lachaux, Lacroix, Rozi{\`e}re, Goyal, Hambro, Azhar et~al.}]{touvron2023llama}
Hugo Touvron, Thibaut Lavril, Gautier Izacard, Xavier Martinet, Marie-Anne Lachaux, Timoth{\'e}e Lacroix, Baptiste Rozi{\`e}re, Naman Goyal, Eric Hambro, Faisal Azhar, et~al. 2023{\natexlab{a}}.
\newblock Llama: Open and efficient foundation language models.
\newblock \emph{arXiv preprint arXiv:2302.13971}.

\bibitem[{Touvron et~al.(2023{\natexlab{b}})Touvron, Martin, Stone, Albert, Almahairi, Babaei, Bashlykov, Batra, Bhargava, Bhosale et~al.}]{touvron2023llama2}
Hugo Touvron, Louis Martin, Kevin Stone, Peter Albert, Amjad Almahairi, Yasmine Babaei, Nikolay Bashlykov, Soumya Batra, Prajjwal Bhargava, Shruti Bhosale, et~al. 2023{\natexlab{b}}.
\newblock Llama 2: Open foundation and fine-tuned chat models.
\newblock \emph{arXiv preprint arXiv:2307.09288}.

\bibitem[{Vo et~al.(2024)Vo, Zhao, Le, Bonilla, and Phung}]{vo2024optimal}
Vy~Vo, He~Zhao, Trung Le, Edwin~V Bonilla, and Dinh Phung. 2024.
\newblock Optimal transport for structure learning under missing data.
\newblock In \emph{International Conference on Machine Learning}.

\bibitem[{Vuong et~al.(2023)Vuong, Le, Zhao, Zheng, Harandi, Cai, and Phung}]{vuong2023vector}
Long~Tung Vuong, Trung Le, He~Zhao, Chuanxia Zheng, Mehrtash Harandi, Jianfei Cai, and Dinh Phung. 2023.
\newblock Vector quantized wasserstein auto-encoder.
\newblock In \emph{International Conference on Machine Learning}, pages 35223--35242. PMLR.

\bibitem[{Wallach et~al.(2009)Wallach, Murray, Salakhutdinov, and Mimno}]{wallach2009evaluation}
Hanna~M Wallach, Iain Murray, Ruslan Salakhutdinov, and David Mimno. 2009.
\newblock Evaluation methods for topic models.
\newblock In \emph{Proceedings of the 26th annual international conference on machine learning}, pages 1105--1112.

\bibitem[{Wang et~al.(2022)Wang, Guo, Zhao, Zheng, Tanwisuth, Chen, Zhou et~al.}]{wanrepresenting2022}
Dongsheng Wang, Dandan Guo, He~Zhao, Huangjie Zheng, Korawat Tanwisuth, Bo~Chen, Mingyuan Zhou, et~al. 2022.
\newblock Representing mixtures of word embeddings with mixtures of topic embeddings.
\newblock In \emph{International Conference on Learning Representations}.

\bibitem[{Wang et~al.(2023)Wang, Liang, Meng, Zou, Li, Qu, and Zhou}]{wang-etal-2023-zero}
Jiaan Wang, Yunlong Liang, Fandong Meng, Beiqi Zou, Zhixu Li, Jianfeng Qu, and Jie Zhou. 2023.
\newblock Zero-shot cross-lingual summarization via large language models.
\newblock In \emph{Proceedings of the 4th New Frontiers in Summarization Workshop}, pages 12--23, Singapore. Association for Computational Linguistics.

\bibitem[{Yang et~al.(2023)Yang, Zhao, Phung, and Du}]{yang2023towards}
Xiaohao Yang, He~Zhao, Dinh Phung, and Lan Du. 2023.
\newblock Towards generalising neural topical representations.
\newblock \emph{arXiv preprint arXiv:2307.12564}.

\bibitem[{Ye et~al.(2024)Ye, Fan, Song, Zheng, Zhao, dan Guo, and Chang}]{ye2024ptarl}
Hangting Ye, Wei Fan, Xiaozhuang Song, Shun Zheng, He~Zhao, Dan dan Guo, and Yi~Chang. 2024.
\newblock Ptarl: Prototype-based tabular representation learning via space calibration.
\newblock In \emph{International Conference on Learning Representations}.

\bibitem[{Yi and Allan(2009)}]{yi2009comparative}
Xing Yi and James Allan. 2009.
\newblock A comparative study of utilizing topic models for information retrieval.
\newblock In \emph{Advances in Information Retrieval: 31th European Conference on IR Research, ECIR 2009, Toulouse, France, April 6-9, 2009. Proceedings 31}, pages 29--41. Springer.

\bibitem[{Young et~al.(2024)Young, Chen, Li, Huang, Zhang, Zhang, Li, Zhu, Chen, Chang et~al.}]{young2024yi}
Alex Young, Bei Chen, Chao Li, Chengen Huang, Ge~Zhang, Guanwei Zhang, Heng Li, Jiangcheng Zhu, Jianqun Chen, Jing Chang, et~al. 2024.
\newblock Yi: Open foundation models by 01. ai.
\newblock \emph{arXiv preprint arXiv:2403.04652}.

\bibitem[{Zhang et~al.(2024)Zhang, Ladhak, Durmus, Liang, McKeown, and Hashimoto}]{zhang2024benchmarking}
Tianyi Zhang, Faisal Ladhak, Esin Durmus, Percy Liang, Kathleen McKeown, and Tatsunori~B Hashimoto. 2024.
\newblock Benchmarking large language models for news summarization.
\newblock \emph{Transactions of the Association for Computational Linguistics}, 12:39--57.

\bibitem[{Zhao et~al.(2018{\natexlab{a}})Zhao, Du, Buntine, and Zhou}]{zhao2018dirichlet}
He~Zhao, Lan Du, Wray Buntine, and Mingyuan Zhou. 2018{\natexlab{a}}.
\newblock Dirichlet belief networks for topic structure learning.
\newblock In \emph{NeurIPS}, pages 7966--7977.

\bibitem[{Zhao et~al.(2018{\natexlab{b}})Zhao, Du, Buntine, and Zhou}]{zhao2018inter}
He~Zhao, Lan Du, Wray Buntine, and Mingyuan Zhou. 2018{\natexlab{b}}.
\newblock Inter and intra topic structure learning with word embeddings.
\newblock In \emph{ICML}, pages 5887--5896.

\bibitem[{Zhao et~al.(2021{\natexlab{a}})Zhao, Phung, Huynh, Jin, Du, and Buntine}]{ijcai2021p638}
He~Zhao, Dinh Phung, Viet Huynh, Yuan Jin, Lan Du, and Wray Buntine. 2021{\natexlab{a}}.
\newblock Topic modelling meets deep neural networks: A survey.
\newblock In \emph{Proceedings of the Thirtieth International Joint Conference on Artificial Intelligence, {IJCAI-21}}, pages 4713--4720. International Joint Conferences on Artificial Intelligence Organization.
\newblock Survey Track.

\bibitem[{Zhao et~al.(2021{\natexlab{b}})Zhao, Phung, Huynh, Le, and Buntine}]{zhao2020neural}
He~Zhao, Dinh Phung, Viet Huynh, Trung Le, and Wray Buntine. 2021{\natexlab{b}}.
\newblock Neural topic model via optimal transport.
\newblock \emph{International Conference on Learning Representations}.

\bibitem[{Zhao et~al.(2021{\natexlab{c}})Zhao, Phung, Huynh, Le, and Buntine}]{zhaoneural2021}
He~Zhao, Dinh Phung, Viet Huynh, Trung Le, and Wray Buntine. 2021{\natexlab{c}}.
\newblock Neural topic model via optimal transport.
\newblock In \emph{International Conference on Learning Representations}.

\bibitem[{Zhao et~al.(2023)Zhao, Sun, Dezfouli, and Bonilla}]{zhao2023transformed}
He~Zhao, Ke~Sun, Amir Dezfouli, and Edwin~V Bonilla. 2023.
\newblock Transformed distribution matching for missing value imputation.
\newblock In \emph{International Conference on Machine Learning}, pages 42159--42186. PMLR.

\bibitem[{Zhou et~al.(2016)Zhou, Cong, and Chen}]{zhou2016augmentable}
Mingyuan Zhou, Yulai Cong, and Bo~Chen. 2016.
\newblock Augmentable gamma belief networks.
\newblock \emph{JMLR}, 17(163):1--44.

\end{thebibliography}
\bibliographystyle{acl_natbib}

\onecolumn

\appendix

\end{document}